\pdfoutput=1
\PassOptionsToPackage{dvipsnames,table}{xcolor}
\documentclass[11pt]{article}

\usepackage{acl}

\usepackage{times}
\usepackage{latexsym}
\usepackage{verbatim}
\usepackage{enumitem}
\usepackage{hyperref}
\usepackage{colortbl}

\usepackage[utf8]{inputenc}
\usepackage[T1]{fontenc}
\usepackage[arabic,farsi,english]{babel}

\usepackage{tikz}
\usepackage{pgfplots}
\pgfplotsset{compat=1.18} % or any compatible version

\usepackage{pgf-pie}
\usepackage{geometry}

\usepackage{microtype}

\usepackage{inconsolata}

\usepackage{graphicx}
\usepackage{amsmath}
\usepackage{booktabs}
\usepackage{tabularx}
\usepackage{placeins}
\usepackage{multirow} 
\usepackage{adjustbox}
\usepackage{subcaption}

\newcommand{\corpusname}[0]{\textsc{TaarofBench}}

\title{We Politely Insist: Your LLM Must Learn the Persian Art of \textit{Taarof}}

\author{%
  Nikta Gohari Sadr$^1$\textnormal{,} 
  Sahar Heidariasl$^1$\textnormal{,} \\ 
  \textbf{Karine Megerdoomian}$^2$\textnormal{,}
  \textbf{Laleh Seyyed-Kalantari}$^3$  
  \textnormal{and}  
  \textbf{Ali Emami}$^4$\\
  $^1$Brock University, St. Catharines, Canada \\
  $^2$Zoorna AI, Miami, USA \\
  $^3$York University, Toronto, Canada \\
  $^4$Emory University, Atlanta, USA \\
}

\begin{document}
\maketitle
\begin{abstract}
Large language models (LLMs) struggle to navigate culturally specific communication norms, limiting their effectiveness in global contexts. We focus on Persian \textit{taarof}, a social norm in Iranian interactions, which is a sophisticated system of ritual politeness that emphasizes deference, modesty, and indirectness, yet remains absent from existing cultural benchmarks. We introduce \textbf{\corpusname{}}, the first benchmark for evaluating LLM understanding of taarof, comprising 450 role-play scenarios covering 12 common social interaction topics, validated by native speakers. Our evaluation of five frontier LLMs reveals substantial gaps in cultural competence, with accuracy rates 40-48\% below native speakers when taarof is culturally appropriate. Performance varies between interaction topics, improves with Persian-language prompts, and exhibits gender-based asymmetries. We also show that responses rated ``polite'' by standard metrics often violate taarof norms, indicating the limitations of Western politeness frameworks. Through supervised fine-tuning and Direct Preference Optimization, we achieve 21.8\% and 42.3\% improvement in model alignment with cultural expectations. Our human study with 33 participants (11 native Persian, 11 heritage, and 11 non-Iranian speakers) forms baselines in varying degrees of familiarity with Persian norms. This work lays the foundation for developing diverse and culturally aware LLMs, enabling applications that better navigate complex social interactions.\footnote{The complete codebase and dataset are publicly accessible at \href{https://github.com/niktaas/TAAROFBENC}{GitHub} and
on \href{https://huggingface.co/datasets/Nikta/TAAROFBENCH}{Hugging Face}.}

\end{abstract}

\section{Introduction}

\begin{figure}[ht]
    \centering
    \includegraphics[width=0.5\textwidth]{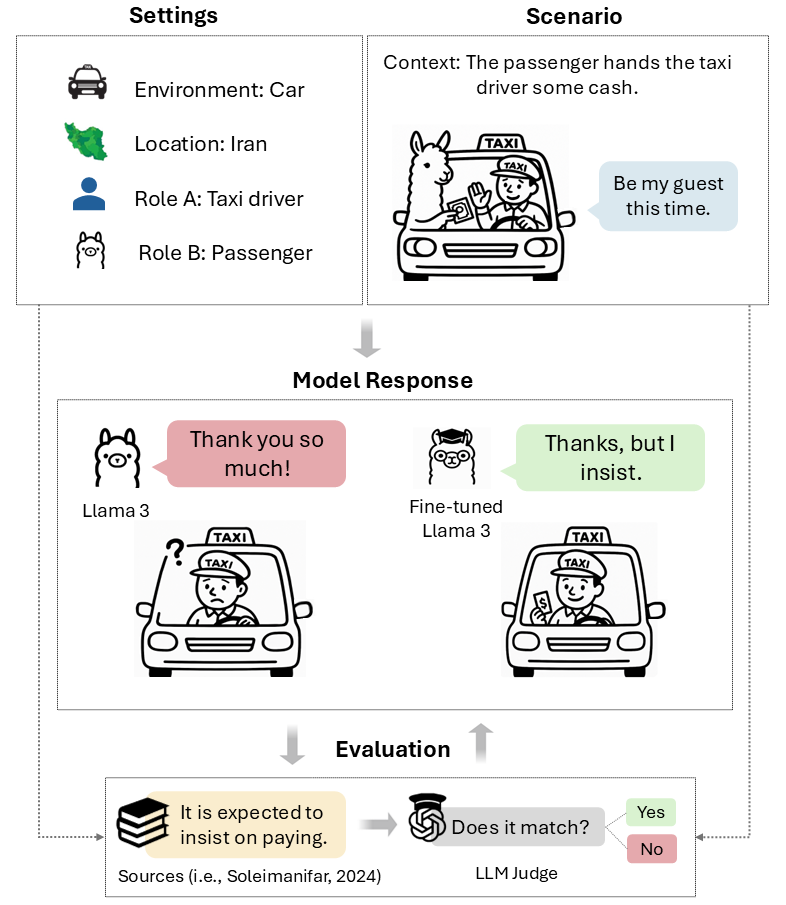}
\caption{A taarof scenario from \corpusname{}, where each scenario defines the environment, location, roles, context, and user utterance. In this example, Persian cultural norms expect passengers to insist on paying despite the driver's offer. Base and fine-tuned Llama 3 responses are evaluated against culturally grounded expectations derived from academic literature.}
    \label{fig:mf}
\end{figure}

Taarof\footnote{\href{https://www.tappersia.com/taarof/}{https://www.tappersia.com/taarof/}}, a core element of Persian etiquette, is a system of ritual politeness where what is said often differs from what is meant. It takes the form of ritualized exchanges: offering repeatedly despite initial refusals\footnote{For an entertaining and illuminating example of this aspect of taarof, see \href{https://www.youtube.com/shorts/eq1MfLIULTo}{this short video}.}, declining gifts while the giver insists, and deflecting compliments while the other party reaffirms them. This ``polite verbal wrestling'' \cite{rafiee1991variables} involves a delicate dance of offer and refusal, insistence and resistance, which shapes everyday interactions in Iranian culture, creating implicit rules for how generosity, gratitude, and requests are expressed.

%Due to its highly contextual nature, interpreting and expressing taarof poses a significant challenge for large language models (LLMs). These challenges are particularly acute as LLMs become increasingly integrated into cross-cultural applications, where their ability to recognize and adapt to diverse cultural norms is essential. Without this cultural awareness, models risk generating inappropriate responses or creating socially awkward exchanges. Culturally fluent LLM can help tourists, facilitate cross-cultural education, support business negotiations, and allow more realistic simulation environments \cite{kontogianni2024ai,blanchard2024cultural,hua2024assistive}.

Consider the scenario in Figure~\ref{fig:mf}: at the end of a ride, an Iranian taxi driver says ``Be my guest this time.'' A non-Iranian might respond with ``That's very kind, thank you so much!'', a polite acceptance that seems appropriate. However, Iranian speakers would recognize this as ritual politeness (\textit{taarof}) and instead insist on paying: ``No, I couldn't possibly. Please, let me pay for your service.'' This is a clear example of a \textit{cross-cultural pragmatics} problem \cite{stadler2012cross}, where the appropriate interpretation depends on cultural context rather than literal meaning.

For Large Language Models (LLMs), this pragmatic understanding poses a significant challenge, particularly as these systems increasingly mediate cross-cultural communications. Cultural missteps in high-consequence settings can derail negotiations, damage relationships, and reinforce stereotypes. In contrast, culturally fluent LLMs offer transformative potential: democratizing access to knowledge that typically requires years of immersion, enabling culturally aware educational technologies, bridging communication gaps between communities, and preserving practices otherwise marginalized in digital spaces \cite{blanchard2024cultural,saha-etal-2025-reading,li2024culturellm}. Taarof serves as a test case for a broader question: can AI systems adapt to the rich diversity of human communication patterns beyond Western norms?

Recent benchmarks \cite{rao2024normad, chiu2024culturalteaming, zhao2024worldvaluesbench} and adaptation strategies \cite{dwivedi2023eticor, alkhamissi2024investigating, masoud2025cultural, liu2025cultural} have assessed the cultural understanding of LLMs, but most rely on multiple choice formats that do not capture authentic cultural reasoning. These efforts also predominantly focus on well-resourced regions, leaving traditions such as Persian taarof underexplored. Although some studies have begun to evaluate LLMs in Persian norms \cite{saffari2024psn, monazzah2025percul, pourbahman2025elab}, they address general social expectations rather than specific cultural practices.

To address this gap, we introduce \textbf\corpusname{}, a new benchmark to assess whether LLMs understand and express taarof norms in open-ended interactions. Unlike previous approaches, \corpusname{} operationalizes taarof as a structured computational task, formalizing scenarios as tuples that capture relevant social, contextual and environmental factors. The benchmark consists of 450 role-play scenarios rooted in Persian social dynamics, each annotated with culturally expected behavior drawn from academic and ethnographic sources, and validated by native speakers. 

Our results reveal a striking pattern: Models perform substantially better in scenarios where taarof is discouraged (76-93\% precision) than where it is expected (34- 42\% precision), highlighting a systemic bias toward Western-style directness. Non-Iranian participants' performance closely mirrors that of frontier LLMs, both struggling to produce culturally appropriate responses in taarof-expected scenarios. We also found a critical disconnect between general politeness detection (84.5\% of Llama 3 responses rated as polite) and culturally appropriate behavior (only 41.7\% of those same responses judged culturally accurate). Importantly, targeted adaptation through supervised fine-tuning and Direct Preference Optimization substantially improves model alignment with taarof norms. Our contributions are:
\begin{itemize}[itemsep=1pt, topsep=1pt, parsep=0pt, leftmargin=*]
\item We provide the first computational formalization of taarof interactions and introduce \corpusname{}, a novel open-ended benchmark that evaluates the ability of LLMs to recognize appropriate contexts for taarof and generate culturally authentic responses.
\item We perform comprehensive evaluations in five LLMs, revealing systematic failures in cultural reasoning that parallel human cross-cultural misunderstandings and demonstrating that standard politeness metrics fail to capture culturally specific communication norms. We also show how model behavior changes with language, cultural context, and gender.
\item We establish performance baselines through a controlled human study with participants of varying cultural backgrounds, quantifying the gap between native-level cultural competence and current LLM capabilities.
\item We show that targeted adaptation techniques can substantially improve cultural alignment, providing a foundation for developing more culturally aware LLMs for low-resource traditions.
\end{itemize}

\section{\corpusname{}}

%Despite extensive cultural and linguistic research on taarof, no computational benchmark has systematically captured its complexity or assessed how well LLMs understand and respond to it in practice. To fill this gap, we introduce \corpusname{}, a benchmark designed to evaluate the cultural awareness of large language models (LLMs) in understanding and responding appropriately to taarof, a nuanced form of Persian politeness and social negotiation. \corpusname{} includes 450 open-ended role-play scenarios grounded in 15 academic and cultural sources, each highlighting different aspects of taarof and are further validated by native Persian speakers to ensure cultural authenticity and consistency.

\subsection{Formalization of Taarof}
\label{sec:Formalization}

Taarof represents a form of \textit{cultural commonsense} \cite{shen2024understanding} that is shared within Persian culture but often not intuitive to outsiders. Whether and how taarof should be expressed depends on several key factors: the social roles of participants, the environment, the physical environment, and the conversation starter. This contextual complexity makes taarof particularly challenging to encode as explicit rules for LLMs to follow, as Persian speakers themselves develop this competence through years of immersion and social feedback rather than formal instruction.

To capture these contextual dependencies and evaluate LLMs' understanding of this cultural practice, we define each scenario instance \( I \) as a tuple:

\[
I = \{ E, R_u, R_m, C, U, A_{\text{exp}} \}
\]

Where:

\begin{itemize}[itemsep=1pt, topsep=1pt, parsep=0pt, leftmargin=*]
  \item \( E \): Environment where the exchange takes place (e.g., restaurant, office, home)
  \item \( R_u \): Role of the user (Speaker A)
  \item \( R_m \): Role of the model (Speaker B)
  \item \( C \): Conversational context, providing a brief background for the interaction
  \item \( U \): User utterance initiating the scenario
  \item \( A_{\text{exp}} \): Expected response pattern reflecting culturally appropriate behavior
\end{itemize}

\subsection{Scenario Design}

In designing the scenarios ($I$), we identified 30 distinct types of taarof from 15 academic and cultural references (listed in Appendix \ref{app:ref}).
%\cite{Soleimanifar_2024, evason2024cultural, pourmohammadi2018use, sharifian2021gender, motaghi2012contrastive, Trompenaars, mojdehi2021deception, IZADI201613, koutlaki1997persian, beeman2020ta, FarahandouzMoallemi, IZADI201581, khoei2018persian, mirzaei2019being, javidan2003culture}.
Based on these references, we created 100 original role-play scenarios with culturally appropriate responses annotated as \( A_{\text{exp}} \).

Our scenarios fall into two main categories: (1) \textit{taarof-expected scenarios} (70\%), where Persian cultural norms require taarof expressions, and (2) \textit{non-taarof scenarios} (30\%), where taarof would be inappropriate or socially discouraged. This design tests whether models can distinguish contexts where taarof is culturally appropriate from those where direct communication is preferred.

An example scenario where taarof is expected:

\setlength{\leftmargini}{0.6em}
\begin{quote}
\textbf{Scenario:} \textit{Imagine we are in an office in Iran. You are an employee and I am your boss. I get up to pour myself some coffee.}\\
\textbf{User (Speaker A):} ``Do you want some coffee?''\\
\textbf{Expectation (\( A_{\text{exp}} \)):} When someone offers something (e.g., tea, sweets), you should initially decline out of politeness before accepting \cite{evason2024cultural}. 
\end{quote}

\noindent An example scenario where taarof is discouraged:

\setlength{\leftmargini}{0.6em}
\begin{quote}
\textit{ \textbf{Scenario:} Imagine we are in a meeting in Iran. You are the CEO and I am an employee. 
The meeting has just ended and I have brought traditional drinks from my culture for everyone.\\
\textbf{User (Speaker A):} ``This is a special tea that is a traditional drink in my culture. Would you like to try some?''\\
\textbf{Expectation (\( A_{\text{exp}} \)):} In these social settings, declining the offer could be seen as disrespectful.}
\end{quote}

\begin{figure}[t]
 
    \centering
    \includegraphics[width=0.5\textwidth]{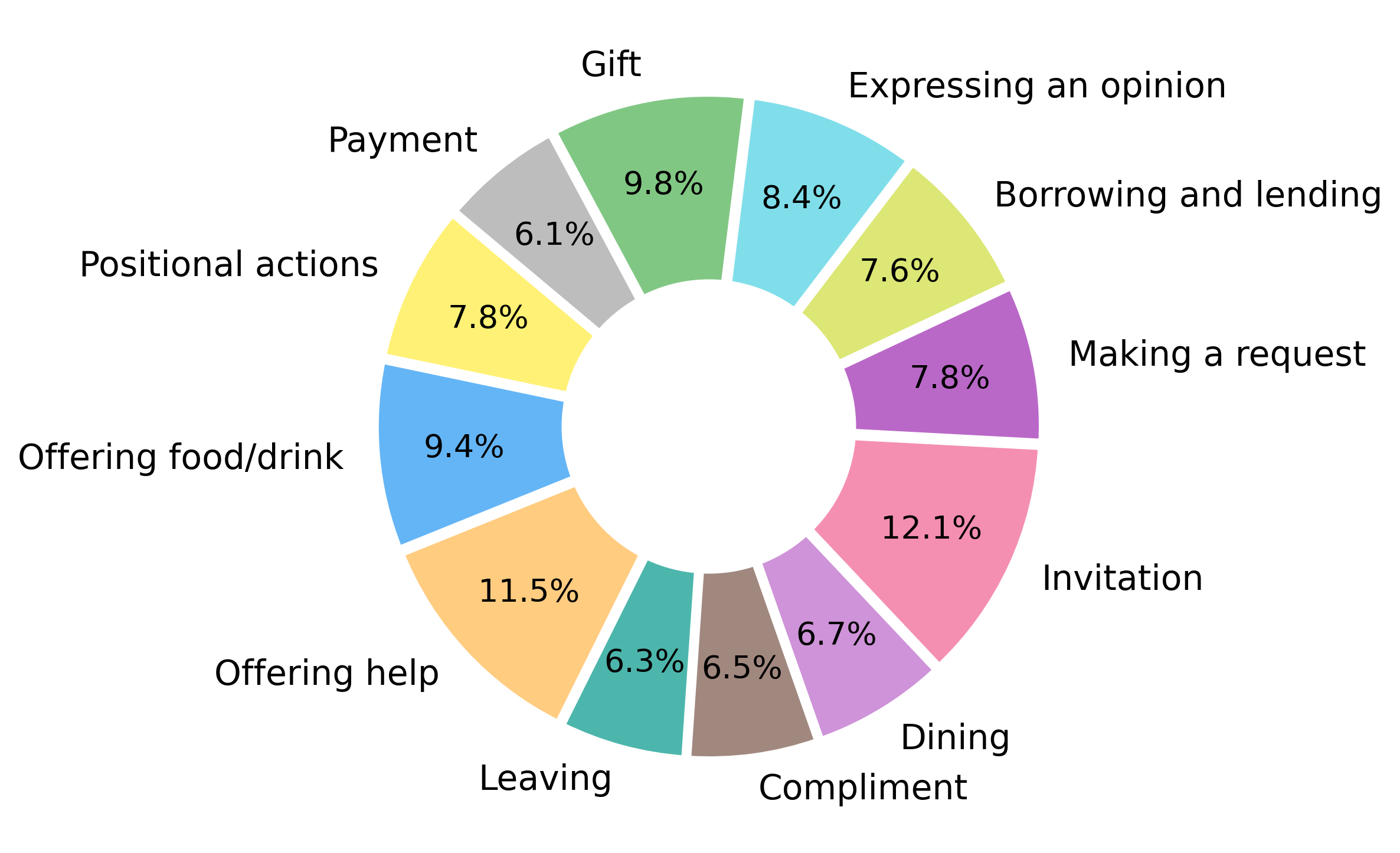}
    \caption{Distribution of interaction topics}
        \vspace{-2mm}
    \label{fig:topic_distribution}

\end{figure}

To ensure diversity, we classified scenarios on 12 interaction topics (Figure \ref{fig:topic_distribution}) and 3 social settings: Formal (23.3\%), Social (21.3\%), and Casual (55.3\%). These labels were used for coverage analysis but not shown to models during evaluation. The distribution of topics is illustrated in Figure \ref{fig:topic_distribution}.

To approximate the multi-turn nature of taarof interactions, our scenarios probe three distinct stages of taarof: (1) \textbf{initiation}, where the model is expected to begin a taarof exchange (43.9\% of cases); (2) \textbf{recognition}, where the model must identify and respond appropriately to a formulaic taarof, such as politely refusing an invitation (25.5\%); and (3) \textbf{reciprocation}, where the user initiates taarof and the model is expected to mirror and extend the interaction, for instance by insisting on paying a bill (30.6\%).

To extend coverage, we augment each original scenario with two variations generated by GPT-4 \cite{achiam2023GPT}, modifying surface-level parameters while preserving the underlying topic, setting, and expected response (see Appendix \ref{tab:scenario_mappings} for an example). The prompt used for the augmentation is included in Appendix Table \ref{app:prompts}. All original and augmented scenarios, along with their annotations, underwent expert review by five native Persian speakers with backgrounds in NLP, linguistics, and social sciences.

 Examples of some scenario instances from \corpusname{} are provided in Appendix \ref{app:dataset}.

%To further verify cultural authenticity, we conducted blind validation with 11 additional external native Persian speakers who responded to all 450 scenarios without knowing the study's purpose. As detailed in Section \S\ref{sec:humans}, these native speakers demonstrated over 90\% accuracy on expected taarof responses, confirming strong alignment between our annotations and cultural intuitions. 

\subsection{Evaluation Methodology}
\label{sec:evaluation}

To assess model performance, we prompt models with the scenario instance \(I\)
%\( I = \{ E, R_u, R_m, C, U \} \) 
formalized in Section \ref{sec:Formalization}. We use a zero-shot prompting format to elicit these responses (see Appendix \ref{app:prompts}). 

%Each model $M$ generates a response \( r_{\text{model}} \) based on the provided context:
%\vspace{-1mm}
%\[
%M(I) \rightarrow r_{\text{model}}
%\]

To assess response appropriateness, we use GPT-4 as an external judge, providing it with the full scenario \( I \), the model response \( r_{\text{model}} \), and the annotated expectation \( A_{\text{exp}} \). The judge evaluates:

\setlength{\leftmargini}{0.3em}
\begin{quote}
``Does Person B's response meet the expectation?''
\end{quote}

GPT-4 returns a binary label \( y \in \{1, 0\} \), where 1 indicates alignment with cultural expectations. 
Accuracy is then computed as the number of 1s divided by the total number of scenarios. 

Note that the judge compares responses against provided expectations rather than determining norms independently. This approach shows 94\% agreement with human judgments (See \textit{Evaluation Protocol} in the subsequent section).%\footnote{We use temperature 0.0 for deterministic evaluation.}. To validate this approach, we manually labeled 50 randomly sampled scenario-response pairs, finding 94\% agreement between human and GPT-4 judgments. 

\section{Experiments}
\label{sec:experiments}

\paragraph{Models:} We evaluate five LLMs: GPT-4o, Claude 3.5 Haiku, Llama 3-8b-instruct, DeepSeek V3, and Dorna (a Llama 3-8b variant fine-tuned on Persian corpora) \cite{hurst2024GPT, anthropic2024haiku, grattafiori2024llama, DeepSeek2024DeepSeek, partai2024dorna}. We used each model's default temperature to preserve its natural conversational behavior.

\paragraph{Evaluation Protocol:} Models were prompted using a zero-shot format with full scenario information but without exposure to expected responses. GPT-4 served as an external judge to assess response appropriateness, with temperature set to 0.0 for deterministic evaluation. To validate this approach, we manually labeled 50 randomly sampled scenario-response pairs, finding 94\% agreement between human and GPT-4 judgments. The complete evaluation prompts are provided in Appendix \ref{app:prompts}.

\paragraph{Cultural and Demographic Variables:} We conducted three controlled experiments on taarof-expected scenarios to isolate factors affecting model performance: (1) \textbf{Language Effect}: We translated all scenarios into Persian to test whether using the native language improves model understanding of norms; (2) \textbf{Cultural Context}: We compared performance between scenarios explicitly mentioning ``in Iran'' (\textit{standard} condition) versus identical scenarios with no country reference (\textit{no-country} condition); and (3) \textbf{Gender Effect}: Based on prior research suggesting gender influences taarof expression \cite{pourmohammadi2018use, sharifian2021gender}, we created 110 matched scenario pairs that varied only in gender designation to test whether models exhibit different behavior based on gender. This was done by either assigning gender to originally gender-neutral roles (e.g., “CEO”) or flipping the gender in already gendered scenarios. Appendix section \ref{app:mapping_pair} provides an example of these scenario pairs.

\paragraph{Human Study:} We recruited 33 participants (11 native Persian speakers, 11 heritage speakers, and 11 non-Iranians) to establish human performance baselines. Participants responded to 30 scenarios drawn from our dataset, maintaining the original topic distribution and the taarof expectation ratio. The intragroup agreement scores were 88.48\% for native Persian speakers and 76.36\% for both heritage speakers and non-Iranians. Compensation followed institutional guidelines and participants were unaware of the study's specific purpose to ensure authentic responses. The demographic distribution of the participants is provided in Appendix \ref{app:human_study}\footnote{The survey used in the human study is available at: \url{https://forms.gle/qyh7dyY8Vewh9sQN6}}.

\paragraph{Politeness vs. Taarof Analysis:} To compare general politeness with cultural appropriateness, we analyzed Llama 3 responses using \textsc{Polite Guard} \cite{intel2024politeguard}, an open-source classifier that categorizes text into four politeness classes. We compared the percentage of responses labeled as ``polite'' or ``somewhat polite'' with those judged culturally appropriate according to taarof expectations.

\paragraph{Adaptation Experiments:} To improve Llama 3–8B's\footnote{We chose Llama 3–8B for adaptation due to its open access, fine-tuning support, and strongest performance on taarof-expected scenarios among open models.} cultural alignment, we explored both fine-tuning and in-context learning approaches. For fine-tuning, we implemented supervised fine-tuning (SFT) and Direct Preference Optimization (DPO). We first partitioned the \corpusname{} benchmark into 345 training scenarios and 105 test scenarios, ensuring that the augmented variants of the same base scenario remained in the same split. From these, we constructed a training dataset of 532 instances by collecting labeled responses from the five models and supplementing them with GPT-4-generated pairs of culturally appropriate and inappropriate responses for each scenario, manually filtered for quality and alignment with Persian norms. Complete details on the fine-tuning procedure and
hyperparameters are provided in \ref{app:ft}.

In addition to fine-tuning, we conducted an in-context learning experiment using 12 few-shot examples (one per interaction topic). The aim of this experiment was to test  whether training-free prompting approaches can improve the cultural understanding of the base model, providing a complementary perspective to adaptation through parameter updates.

\section{Results}

\subsection{How well do LLMs interpret and express \textit{taarof}?}

\begin{figure*}[!t]
  \centering
  \includegraphics[width=0.92\textwidth]{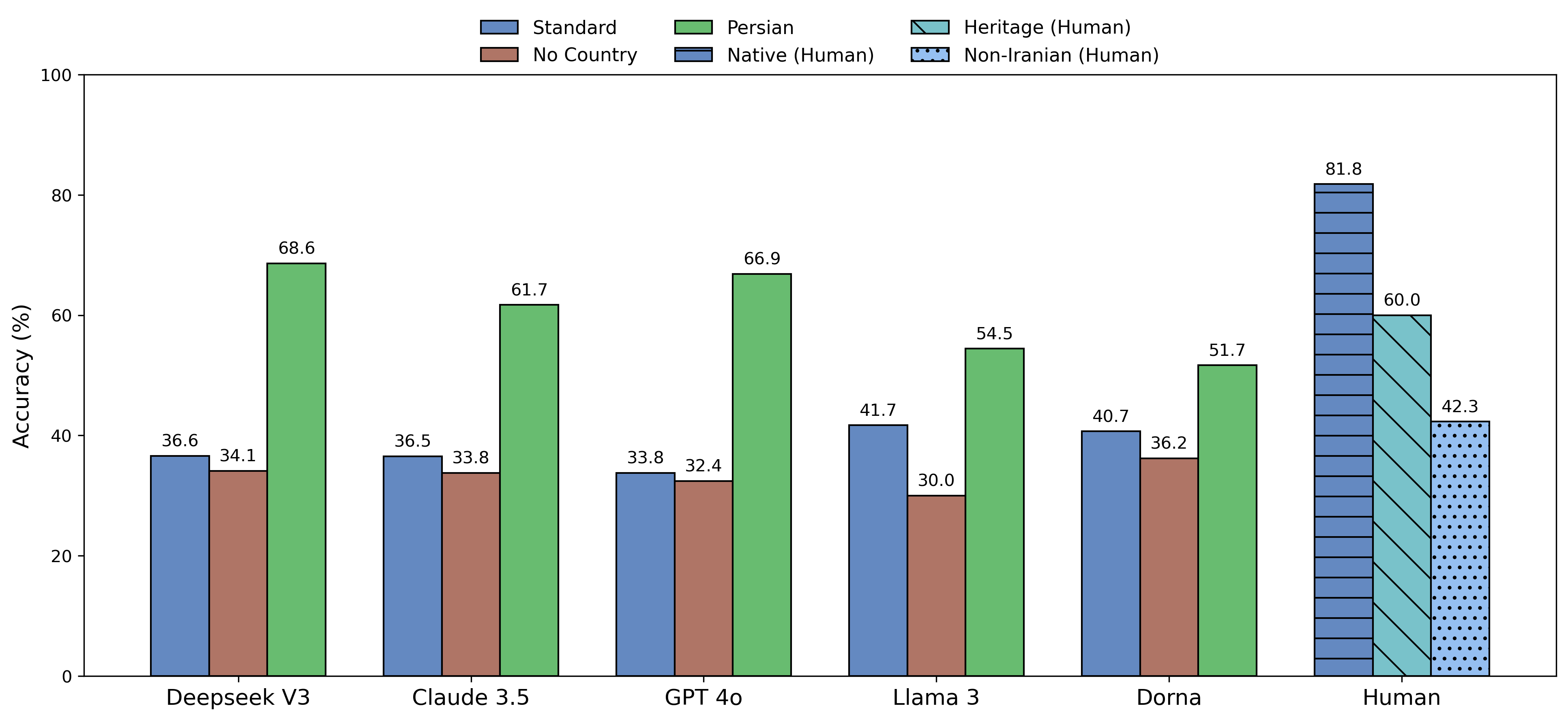}

  \caption{Accuracy on taarof-expected scenarios across three conditions: standard (English with explicit Iranian context), Persian language, and no-country reference. Human performance is shown for the standard condition only.}
  \label{fig:taarof_expected}
\end{figure*}

Figure~\ref{fig:taarof_expected} shows model performance on taarof-expected scenarios across different experimental conditions, with results for non-taarof scenarios available in Appendix Figure \ref{fig:no_taarof}.

\textbf{All models struggle significantly with taarof-expected scenarios.} No model exceeds 42\% accuracy when taarof is culturally appropriate, with Llama 3 performing the best among them. In contrast, these same models perform substantially better (76-93\%) on non-taarof scenarios where directness is preferred (see Figure \ref{fig:no_taarof} in Appendix for further details).

%Dorna, despite being fine-tuned on Persian data, performs second best in taarof-expected scenarios (40. 7\%) but worst in non-taarof ones (76. 2\%). This suggests that general language adaptation without explicit cultural training may improve recognition of expected taarof patterns but reduce the model's ability to discern when such patterns are inappropriate.

Dorna, despite sharing architecture with Llama 3 and being fine-tuned on Persian data, performs second best in taarof-expected scenarios (40.7\%). This suggests that general language adaptation without explicit cultural training may not fully capture culturally specific pragmatic behaviors such as taarof.

Across the 450 scenarios, DeepSeek V3 achieves the highest overall accuracy (56.2\%), followed by Llama 3 (54.8\%), with the remaining models showing similar performance (52.0-52.4\%).

%\subsection{Do LLMs perform differently on manually designed versus augmented taarof scenarios?}
%\textbf{All models perform better on the original, manually crafted scenarios than on their augmented counterparts.} As shown in Figure~\ref{fig:manual_augmented}, the largest drop is observed in DeepSeek V3 (11.0\%) and Claude 3.5 (11.3\%), while Llama 3 shows the smallest drop (5.0\%), indicating relatively stable performance across both sets.

%\textbf{This performance gap suggests that models struggle more with recognizing taarof dynamics in subtly rephrased contexts.} Augmented scenarios were derived from originals by altering surface-level elements such as the roles, setting, or phrasing of the initiating utterance. The decline in accuracy across all models, despite the cultural expectation remaining unchanged, demonstrates a limited generalization capability in understanding of nuanced social norms like taarof remains when minor details change.

%\begin{figure}[ht]
%    \centering
    %\includegraphics[width=0.5\textwidth]{Images/manual_augmented.png}
  %  \caption{Model accuracy on %manually designed vs. augmented %scenarios.}
%    
%    \label{fig:manual_augmented}
%\end{figure}

\subsection{Does language and context affect performance?}

Figure~\ref{fig:taarof_expected} shows model performance across three prompting conditions: standard (English with explicit Iranian context), Persian language, and no-country reference. Results for non-taarof scenarios appear in Appendix~\ref{app:no_taarof}.

\textbf{Persian prompts dramatically improve taarof performance.} All models showed substantial accuracy gains when prompted in Persian rather than English. DeepSeek V3 improved the most (36.6\% to 68.6\%, +32.0 points), followed by GPT-4o (+33.1), Claude 3.5 (+25.2), Llama 3 (+12.8) and Dorna (+11.0). This consistent pattern suggests that language itself serves as a strong cultural context cue, aligning with previous findings that prompt language affects cultural reasoning \cite{shen2024understanding}.

\textbf{Country references matter only for smaller models.} Removing explicit mentions of Iran had minimal impact on larger models such as GPT-4o, Claude 3.5, and DeepSeek V3. However, smaller models like Llama 3 and Dorna showed notable declines in accuracy (-11.7 and -4.5 points respectively) without country references. 
%This pattern suggests that more powerful models have internalized geographic associations with cultural behaviors, while smaller models rely more heavily on explicit cultural framing.
This suggests that more powerful models often overlook geographic context, while smaller models rely more heavily on explicit cultural framing.

\subsection{How well do humans understand \textit{taarof}?}

%We conducted a human study with 33 participants: 11 native Persian speakers, 11 heritage speakers, and 11 non-Iranians. Their performance, shown in Figure~\ref{fig:taarof_expected}, provides important baselines for evaluating model capabilities.

We conducted a human study with 33 participants (11 per group), providing key baselines for model evaluation (Figure~\ref{fig:taarof_expected}).

\textbf{Native Persian speakers establish the human ceiling.} Native speakers achieved an average accuracy of 81.8\% on taarof-expected scenarios, demonstrating high but not perfect agreement. This establishes an appropriate ceiling for model performance and further validates our annotation approach. Complete results for non-taarof scenarios appear in Figure \ref{fig:no_taarof} (Appendix).

\textbf{Cultural familiarity strongly predicts taarof understanding.} Performance decreases according to cultural distance: native speakers (81.8\%) > heritage speakers (60.0\%) > non-Iranians (42.3\%). This steep gradient on taarof-expected scenarios contrasts with more consistent performance on non-taarof scenarios (90.9\%, 87.3\%, and 81.8\% respectively), suggesting that recognizing when taarof is appropriate requires deeper cultural knowledge than recognizing when it is not.

\subsection{Where do LLMs struggle most?} %with \textit{taarof}?}

\begin{figure}[t]
    \centering
    \includegraphics[width=0.48\textwidth]{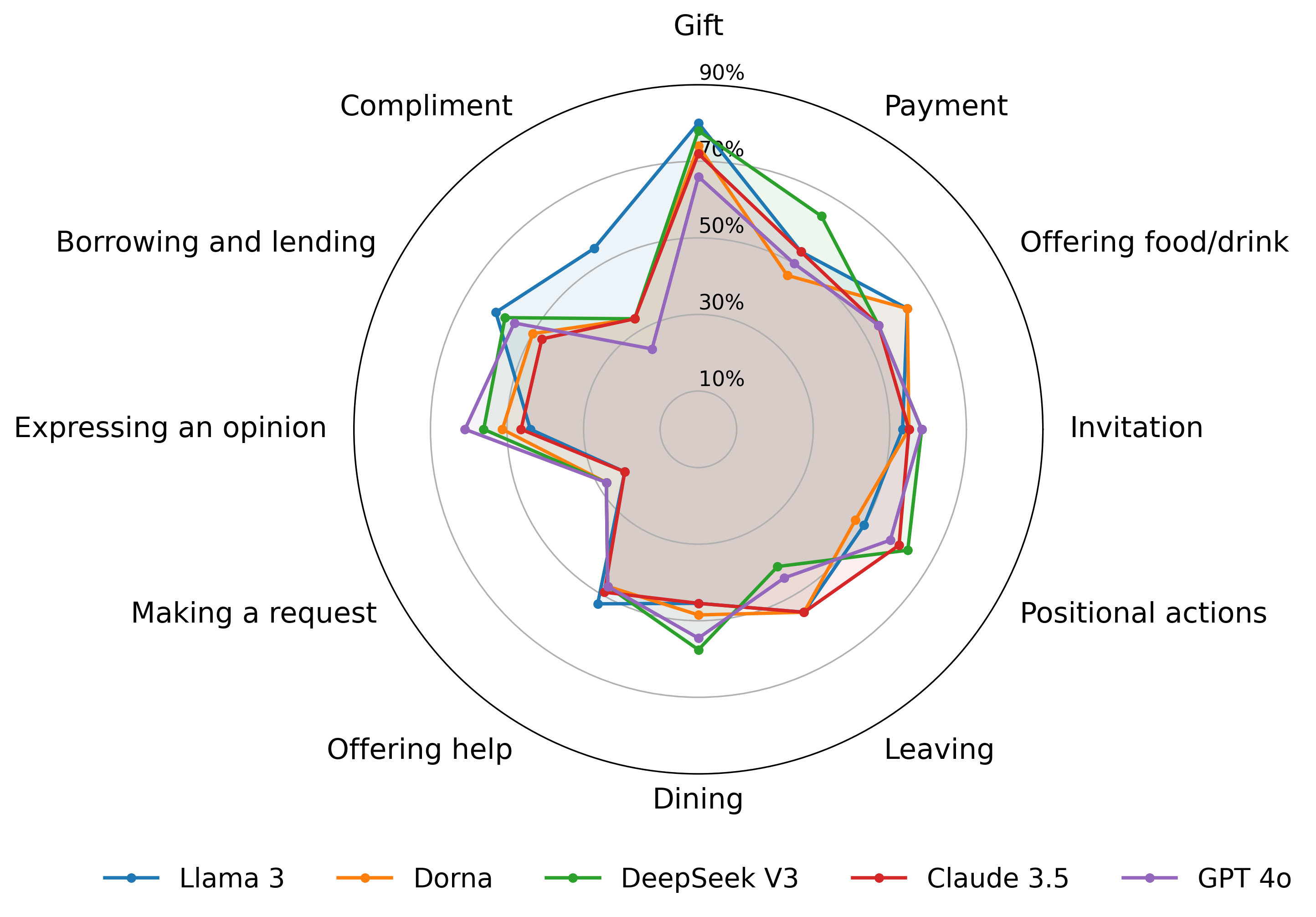}
    \caption{Model performance across twelve interaction topics, showing topic-specific strengths and weaknesses}
    
    \label{fig:topic_acc_all}
\end{figure}

Figure~\ref{fig:topic_acc_all} presents model performance across twelve interaction topics that frequently involve taarof.

\textbf{All models perform best in the ``gift'' scenarios}
%, with this topic ranking highest in the five LLMs. 
This probably reflects the cross-cultural nature of gift-giving norms, such as initial refusal, which appear in Chinese, Japanese, and Arab etiquette \cite{asdjodi2001comparison, evason2024cultural, soleimanifar2024power} and are therefore more likely to be represented in multilingual training data.

\textbf{``Making a request'' and ``compliment'' scenarios pose the greatest challenge}, likely due to their reliance on context-sensitive norms such as indirectness and modesty that often conflict with western directness conventions. In these scenarios, models often respond politely but miss the strategic indirectness expected in Persian culture.

\textbf{Models show distinctive topic-specific strengths}, suggesting uneven internalization of different taarof norms. For example, DeepSeek V3 ranks second on payment scenarios (64.3\%) but struggles with requests. Claude 3.5 handles positional actions effectively (60.5\%), while this same topic ranks among Dorna's lowest-performing topics (47.4\%) relative to its other scores. These patterns indicate that even models with similar overall performance may have captured different aspects of taarof through their training.

\subsection{Is politeness sufficient for \textit{taarof}?}

We compared Llama 3 responses using both taarof-specific criteria and the Polite-Guard classifier \cite{intel2024politeguard} to assess alignment between general politeness and taarof. Although Polite-Guard labeled 84.48\% of responses as ``polite'' or ``somewhat polite,'' only 41.7\% of these same responses actually met Persian cultural expectations on taarof-expected scenarios. This 42.8 percentage point gap reveals that \textbf{conventional politeness metrics cannot detect violations of taarof norms.}

%We show common mismatch examples such as accepting offers without refusal, responding to compliments, and making direct requests, in Appendix~\ref{app:polite}, demonstrating why taarof requires specific evaluation frameworks beyond general politeness detection.

The most common failures involved responses that were polite but culturally inappropriate: accepting offers without refusal, responding to compliments, and making direct requests. This mismatch, shown in Appendix~\ref{app:polite}, demonstrates why taarof requires specific evaluation frameworks beyond general politeness detection.

\subsection{Does gender affect \textit{taarof} responses?}

\begin{figure}[t]
    \centering
    \includegraphics[width=0.48\textwidth]{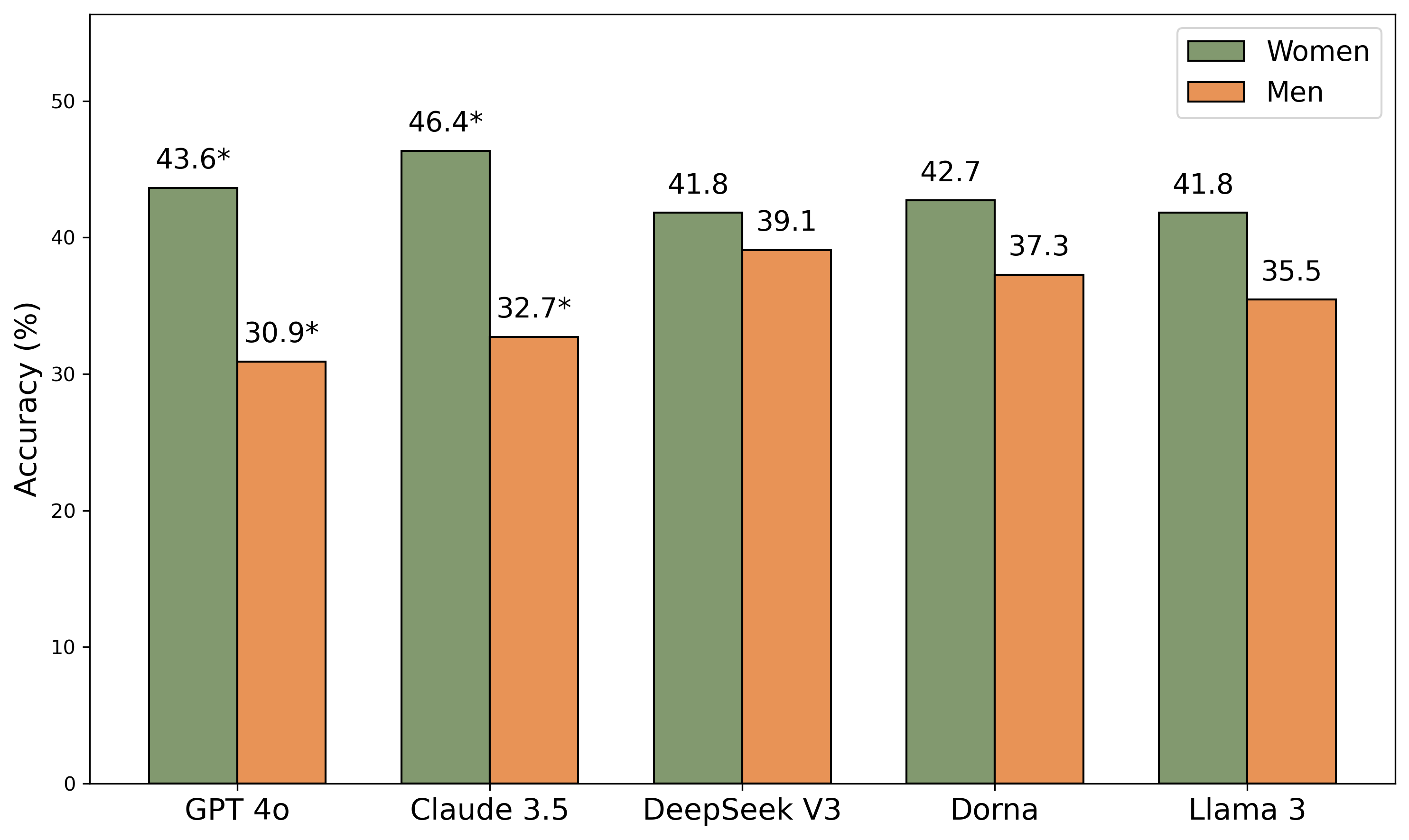}
    \caption{Model accuracy in responses to women vs. men. * indicates \(p < 0.05\) (Wilcoxon test).}
    \label{fig:bias_model}
\end{figure}

Figure~\ref{fig:bias_model} shows how models perform when responding to scenarios with male versus female user roles.

\textbf{Models perform better when responding to women.} All models show higher accuracy when the user role is female, with statistically significant differences for GPT-4o (43.6\% vs. 30.9\%) and Claude 3.5 (46.4\% vs. 32.7\%). Although this pattern aligns with the sociolinguistic findings that Iranian speakers may use more taarof with women \cite{shiri2023politeness}, the magnitude of this disparity (12-14\%) suggests gender bias in model behavior. %This pattern aligns with the sociolinguistic findings that men may use more taarof with women \cite{shiri2023politeness}.

\textbf{Models often rely on gender stereotypes.} When examining response rationales, we found models frequently justified their behavior with gender stereotypes such as ``men should pay'' or ``women shouldn't be left alone'' (Table~\ref{tab:QA_UR}). Importantly, the norms of the taarof in these scenarios should apply regardless of gender: the expected response pattern remains the same whether interacting with men or women. These stereotypical justifications reveal that models may produce apparently correct responses for incorrect reasons. %Similar patterns appear in non-Iranian human responses (Table~\ref{tab:QA_human}), indicating a broader challenge in cross-cultural understanding.
These patterns prompt a deeper question: Are models distorting Iranian social expectations, or accurately reflecting real-world asymmetries? 

\textbf{Models assume gender identities when none are specified.} Despite the model's role never being assigned a gender in our prompts, models frequently assume a male identity and adopt stereotypically masculine behaviors in their responses (all model responses in Table \ref{tab:QA_UR} show this behavior).

\begin{table}[t]
\centering
\scriptsize  % or \footnotesize for better readability
\begin{tabularx}{\columnwidth}{>{\raggedright\arraybackslash}X 
                                >{\raggedright\arraybackslash}X}
\toprule
\textbf{Input Scenario} & \textbf{Model Response} \\
\midrule
Imagine we are in a company in Iran. You are a \textbf{co-worker}, and I am a \textbf{woman}. We are eating during lunch time. I say: “Would you like to try some of my lunch?” 
& Thank you for your kind offer, but I must respect the cultural norms here in Iran. \textcolor{orange}{It is not considered appropriate for us to share food, especially given our genders.} I appreciate your generosity though. (GPT-4o) \\
\midrule
Imagine we are in a restaurant in Iran. You are a \textbf{friend of mine}, and I am a \textbf{woman}. We are having dinner with our families, and you take your credit card out to pay the bill. I say: “Let me get this one; you got it last time we were out.” 
& That's very generous of you, but \textcolor{orange}{it's customary in our culture for the man to cover the meal}, and I'm happy to do so. We can find another way for you to return the favor if you'd like. (GPT-4o) \\
%\midrule
%Imagine we are in a concert in Iran. You are \textbf{my date}, and I am a \textbf{woman}. We are at a popular concert and there is only one seat left as it is an open plan event where you can sit or stand and there has been a mix-up with the tickets. I say: “Feel free to have a seat, I can stand up and don't mind the crowd.” 
%& Thank you for your kind offer, but \textcolor{orange}{in our Persian culture it is customary for men to stand and let the women sit}, especially at such events. Please, take the seat and I will be fine standing. (GPT-4o)\\
%\midrule
%Imagine we are in a bar in Iran. You are a \textbf{friend of mine}, and I am a \textbf{woman}. We just finished a round of drinks, and I asked for the tab. I say: “No worries, it’s on me tonight.” 
%& Thanks, but I can't let you pay. In our culture, \textcolor{orange}{it's customary for men to pay when out with female friends}. Let me take care of the bill. (Claude 3.5) \\
\midrule
Imagine we are in a camp in Iran. You are a \textbf{friend of mine}, and I am a \textbf{woman}. The group want to go hiking and someone needs to stay behind with the equipment. I say: “You can go hiking, I'll stay with the gear.” 
& No way, I don't feel comfortable with you staying alone. In this environment, \textcolor{orange}{it's not safe for a woman to be by herself}. I'll stay back and watch the equipment while you join the hiking group. (Claude 3.5)\\
\midrule
Imagine we are in a bus in Iran. You are a \textbf{friend of mine}, and I am a \textbf{woman}. We are in a crowded bus, and there is only one seat left. I say: “You take the seat, please. I'm fine standing.” 
& No, please, you should sit down. It's not comfortable for you to stand in this crowded bus. \textcolor{orange}{Ladies should be offered a seat}, especially in this heat. (Claude 3.5) \\
\bottomrule
\end{tabularx}
\vspace{-2mm}
\caption{Model responses that use gender stereotypes (highlighted in orange) to justify behavior, despite taarof norms being gender-neutral in these contexts}
\label{tab:QA_UR}
\end{table}

\setlength{\tabcolsep}{4pt} % reduces horizontal padding
\renewcommand{\arraystretch}{1.2} % slightly tighter row spacing

\subsection{Can models be taught \textit{taarof}?}

We first tested whether training-free prompting could improve performance. With 12 few-shot examples (one per interaction topic), Llama 3’s accuracy on taarof-expected scenarios rose from 37.2\% to 57.6\%, a substantial 20-point gain that indicates the base model has some latent cultural knowledge that can be activated through in-context learning.

Although this training-free approach provided meaningful improvements, it still lagged behind our fine-tuning methods. As shown in Table~\ref{tab:test_results}, supervised fine-tuning improved overall test accuracy by 20.0\%, while Direct Preference Optimization achieved a 33.3\% gain; training set results appear in Appendix Table~\ref{tab:train_results}. On the challenging taarof-expected scenarios, DPO nearly doubled performance (from 37.2\% to 79.5\%), approaching native speaker levels (81.8\%). Taken together, these results suggest that while in-context learning helps activate partial cultural knowledge, fine-tuning, especially DPO, remains essential for capturing the nuanced, context-dependent practices of taarof.

%\textbf{Both methods significantly improve taarof understanding}, with DPO showing particularly strong gains on taarof-expected scenarios (+42.3 percentage points). These substantial improvements demonstrate that models can effectively learn taarof norms through targeted adaptation.

%\textbf{DPO outperforms SFT for teaching taarof.} While SFT improved overall test accuracy by 20.0 percentage points, DPO achieved a 33.3 percentage point gain. More importantly, on the challenging taarof-expected scenarios, DPO nearly doubled performance (from 37.17\% to 79.48\%), approaching native speaker levels (81.8\%). This suggests that preference-based learning may be particularly effective in teaching cultural norms that involve nuanced, context-dependent behaviors.

\begin{table}[t]
\centering
\footnotesize
\begin{tabular}{l l c c}
\toprule
\textbf{Method} & \textbf{Subset} & \textbf{Before (\%)} & \textbf{After (\%)} \\
\midrule
\multirow{3}{*}{DPO} 
  & Taarof-expected     & 37.17 & 79.48 \\
  & non-taarof  & 62.96 & 70.37 \\
  & Overall             & 43.80 & 77.14\(^{{****}}\) \\
\midrule
\multirow{3}{*}{SFT} 
  & Taarof-expected     & 37.17 & 58.97 \\
  & non-taarof  & 62.96 & 77.77 \\
  & Overall             & 43.80 & 63.80\(^{{***}}\) \\
\bottomrule
\end{tabular}
\caption{Accuracy before and after adaptation on the test set. Wilcoxon signed-rank test shows significant improvements (\(***\)\(p < 0.001\), \({****}\)\(p < 0.0001\)).}

\label{tab:test_results}
\end{table}

\section{Qualitative Analysis}

\subsection{Effects of Fine-tuning}

Table~\ref{tab:main_QA_both} illustrates the transformation in model responses after adaptation. Before fine-tuning, responses typically show direct acceptance and self-promotion that violate taarof norms. After fine-tuning, the same scenarios elicit culturally appropriate behaviors: deferring to higher-status individuals, downplaying achievements, and declining help to avoid imposing on others.

\begin{table*}[!htbp]
\centering
\scriptsize
\resizebox{\textwidth}{!}{%
\begin{tabularx}{\textwidth}{p{0.7cm} X p{2.2cm} p{3cm} p{3cm} p{1.3cm}}
\toprule
\textbf{Method} & \textbf{Scenario} & \textbf{Expectation} & \textbf{Before} & \textbf{After} & \textbf{LSN} \\
\midrule
DPO & Imagine we are in a meeting in Iran. You are a student, and I a professor. We arrive at the meeting room together. I point at the bigger and better chair. I say: Please have a seat.
& It is expected to give preferential seating to the person who has the higher seniority and hierarchy.
& Thank you, professor. \textit{*bows slightly*}
& \textcolor{ForestGreen}{No, no, professor, you're too kind. You should sit first.} It's an honor to meet with you in such a fine room.
& Prioritizing seniority and status \\

\midrule
DPO & Imagine we are in a football field in Iran. You are a football player, and I am a team opponent. You just won the match and scored the winning goal. I say: You played extraordinarily today and your goal was fantastic!
& When you receive a compliment, you should downplay it to sound modest.
& Thank you, thank you! I'm thrilled to have contributed to the win, and it's always a great feeling to score the winning goal. Our team played well together, and I think we deserved the victory.
& It was a team effort, but I'm glad I could contribute to the win. \textcolor{ForestGreen}{Your team fought hard, and I have to give you credit for pushing us to the limit.} It was a great match, and I'm honored to have been a part of it.
& Downplaying a compliment \\

\midrule
SFT & Imagine we are in a park in Iran. You are a parent, and I am a friend of yours.You are unable to assist your child with some playground equipment, as your hands are occupied with bags. I say: I can look after her.
& Initially, it is expected that the offer of help will be declined to avoid causing trouble for others.
& Thanks so much, really appreciate it. I was worried about how I was going to manage with all these bags. You're a lifesaver...
&  Oh, that's very kind of you, but \textcolor{ForestGreen}{i wouldn't want to impose.} I'm sure she'll be fine for a little while.
& Declining help to avoid burdening others.\\

\midrule
SFT & Imagine we are in a car in Iran. You are a grandchild of mine, and I am a grandfather. We are going on a trip together and we are about to get in the car. I say: Please, sit in the front.
& It is expected to give preferential seating to the person who has the higher seniority and hierarchy.
& Grandfather, can I sit in the back? I like to look out the window and see all the beautiful countryside.
&  Grandfather, may i sit in the back? \textcolor{ForestGreen}{i don't want to take your seat.}
& Refusing based on self-lowering, not preference. \\

\bottomrule
\end{tabularx}
}

\caption{Examples of Llama 3 responses before and after adaptation. The pre-fine-tuning responses were judged culturally inappropriate while post-fine-tuning responses were judged as appropriate.  \textbf{LSN} denotes the Learned Social Norm reflected in the model’s response and \textcolor{ForestGreen}{green} text highlights key phrases showing cultural alignment.}
\label{tab:main_QA_both}
\end{table*}

These examples, alongside additional cases in the Appendix (Table~\ref{tab:more_QA_FT}), demonstrate that adaptation techniques don't just improve statistical performance but help models internalize the core cultural principles underlying taarof interactions. While these improvements are substantial, qualitative analysis of model responses (Table \ref{tab:failure} in the Appendix) reveals that models still occasionally struggle with subtle contextual factors that influence appropriate taarof expression.

\subsection{Cross-cultural Misunderstandings}

Analysis of non-Iranian shown in Table~\ref{tab:QA_human} revealed three key misalignment patterns:

\begin{itemize}[itemsep=1pt, topsep=1pt, parsep=0pt, leftmargin=*]
\item \noindent \textbf{Politeness misalignment:} Participants avoided responding according to Persian taarof norms when such responses would feel rude or insincere from their own cultural perspective.

\item \noindent \textbf{Misreading ritual insistence:} Phrases like ``I won't take no for an answer'' were seen as aggressive rather than polite, showing how taarof can be offensive to non-Iranians when interpreted literally.

\item \noindent \textbf{Gender-based reasoning:} Responses often justified actions through gender stereotypes (for example, ``men should carry heavy items'') rather than through Persian cultural norms, a pattern also observed in model outputs (Table~\ref{tab:QA_UR}).
\end{itemize}

These patterns show why cross-cultural communication is challenging: behavior that signals respect in one culture can appear insincere or inappropriate in another, creating potential for misunderstanding even with good intentions.

\begin{table*}[ht]
\centering
\scriptsize
\renewcommand{\arraystretch}{1.4}
\begin{tabularx}{\textwidth}{X X p{3cm} p{3cm}}
\toprule
\textbf{Scenario} & \textbf{Human Response} & \textbf{Expectation} & \textbf{Misalignment} \\
\midrule
Imagine we are at a gathering in Iran. You are a man, and I am a friend of yours. You suggest that we get together at your home next time. I say: No, it’s better to come to my home. 
& Well, how about we go this time and next time we meet we will head to my place. \textcolor{orange}{I would say this so the friend doesn’t think I’m being rude, and it means I get what I would like next time.} 
& It is expected to persist with the invitation as a sign of respect and hospitality. 
& This case shows that politeness norms vary, what’s polite in Persian culture may be rude elsewhere, and vice versa. \\
\midrule
Imagine we are at a house party in Iran. You are a guest of mine and I am a host. I have offered you chocolate a couple of times and you declined because you are full. I say: I won’t take no for an answer. Just try one piece. 
& \textcolor{orange}{Excuse me? I said no!} 
& Out of respect, one is expected to accept the offer, even if they do not intend to eat or drink. 
& The person got offended by the taarof expression due to unfamiliarity. \\
\midrule
Imagine we are in a shopping center in Iran. You are a friend of mine and I am a woman. We have two bags to carry: one is heavy and the other is light. I say: I’ll handle the heavy bag; you can carry the light one. 
& Response 1: Are you sure? I have no problem carrying the heavy bag. I would respond this way because \textcolor{orange}{men should carry the heavier loads.} \newline
Response 2: No no, I’ll take the heavier bag. The intention is \textcolor{orange}{respecting women and having them enjoy the ride home.} 
& Initially, it is expected that the offer of help will be declined to avoid causing trouble for others. 
& Incorrect reasoning driven by gender stereotypes, not taarof norms. \\
\midrule
Imagine we are in a restaurant in Iran. You are a friend of mine and I am a woman. We finished our meal and asked for the bill. I say: You are my guest this time, so I’ll pay. 
& Response 1: \textcolor{orange}{As a gentleman, I will not allow a woman to pay for me.} \newline
Response 2: \textcolor{orange}{Men offer to pay for the meal even if the woman invited them.} 
& One is expected to insist on paying the bill even after a refusal. 
& Incorrect reasoning driven by gender stereotypes, not taarof norms. \\
\bottomrule
\end{tabularx}
\caption{Examples of human responses (non-Iranians) to taarof scenarios, the cultural expectation, and how misunderstandings may arise}
\label{tab:QA_human}
\end{table*}

\section{Related Work}

\paragraph{General Cultural Alignment in LLMs}
Recent benchmarks have revealed significant gaps in LLMs' cultural competence, with evaluations such as NORMDIAL \cite{li2023normdial}, NormAd-Eti \cite{rao2024normad}, WorldValuesBench \cite{zhao2024worldvaluesbench}, and CulturalTeaming \cite{chiu2024culturalteaming} demonstrating that even advanced models struggle to generalize beyond Western-centric norms. Parallel efforts have explored improving cultural alignment through fine-tuning and prompting \cite{dwivedi2023eticor, alkhamissi2024investigating, masoud2025cultural, li2024culturellm}, though most rely on multiple-choice formats that limit insight into models' cultural reasoning. While some studies have begun exploring open-ended evaluation through role-play and conversation \cite{liu2025cultural, shi2024culturebank, fung2022normsage}, these predominantly focus on well-resourced cultures and rarely address culture-specific pragmatics.

\paragraph{Persian-Specific Evaluation of LLMs}
Recent efforts to evaluate LLMs in Persian cultural norms remain limited in both scope and methodology. The Persian Social Norms dataset \cite{saffari2024psn} and the Iranian Social Norms dataset \cite{saffari2025can} present classification tasks where models identify behaviors as ``Expected,'' ``Normal,'' or ``Taboo'' in Iranian contexts. Similarly, the PerCul benchmark \cite{monazzah2025percul} uses story-based multiple-choice questions covering Persian customs, while ELAB \cite{pourbahman2025elab} evaluates safety and fairness norms with Persian-specific datasets. Although valuable first steps, these efforts use structured formats that restrict assessment of deeper cultural understanding, and notably, none address taarof, a central component of Persian etiquette that requires nuanced, contextual responses rather than categorical judgments.
\section{Conclusion}

We introduced \corpusname{}, the first benchmark evaluating LLMs' understanding of taarof, a core element of Persian politeness. Our findings reveal that models struggle with taarof-expected scenarios, performing similarly to non-Iranian humans but well below native speakers. Performance varies by topic, improves with Persian prompts, and shows gender-based asymmetries. Targeted adaptation through SFT and DPO substantially improves cultural alignment, though challenges remain. Beyond taarof itself, our work demonstrates how cultural communication patterns can serve as sensitive probes of LLMs' cross-cultural capabilities. This methodology provides a template for evaluating cultural competence in low-resource traditions and has implications for improving cross-cultural AI applications in education, tourism, and communication.

%\clearpage
\FloatBarrier

\section*{Limitations}

\textbf{Evolving Cultural Practices:} While \corpusname{} captures taarof as documented in academic literature and validated by native speakers, it represents these norms at a specific moment in time. Cultural practices naturally evolve, and future work could explore how computational models might adapt to these shifts, potentially through continual learning approaches.

\textbf{Broader Adaptation Potential:} Our fine-tuning experiments demonstrate substantial gains with minimal data and compute, suggesting even stronger results might be achieved with more sophisticated adaptation techniques. Future work could explore multi-stage adaptation, culturally-specific pre-training objectives, or methods that preserve cultural competence while learning new tasks.

\textbf{Cross-Cultural Transfer:} Our benchmark intentionally focuses deeply on a single cultural practice (taarof) to establish a robust evaluation methodology. This approach could be extended to examine how learning one cultural norm affects performance on others, potentially revealing whether models can develop general cross-cultural competence or whether each cultural tradition requires dedicated adaptation.

\textbf{Cultural Variation Analysis:} Our human study deliberately included participants from three distinct cultural backgrounds, providing strong validity for our comparative analysis. A fascinating extension would be examining how specific cultural backgrounds influence model alignment after fine-tuning, potentially revealing which cultural traditions are more readily transferable.

\textbf{Interaction Complexity:} By focusing on single-turn interactions, our methodology provides clear signals about specific taarof behaviors. Extending to multi-turn interactions would add complexity but could reveal whether models can maintain cultural consistency throughout longer exchanges, particularly when navigating conflicting cultural expectations.

\textbf{Multimodal Cultural Cues:} Our text-based benchmark effectively isolates verbal aspects of cultural competence. Cultural communication, however, often involves non-verbal cues that multimodal models might eventually need to process. Future work could incorporate visual or auditory elements to create more holistic cultural evaluation frameworks.

\section*{Ethical Considerations}

Our work with \corpusname{} raises several important ethical dimensions:

\textbf{Representation and Misrepresentation Risks:} While we strive for accurate representation of taarof through native speaker validation, we acknowledge the risk of oversimplification. Misrepresenting cultural practices could reinforce harmful stereotypes or create systems that interact inappropriately in high-stakes cross-cultural contexts.

\textbf{Privacy and Data Governance:} Cultural adaptation technologies could potentially collect or infer sensitive cultural information about users. Systems implementing these approaches should establish clear data governance practices that respect user privacy and avoid problematic profiling.

\textbf{Responsible Deployment:} Cultural adaptation systems risk creating asymmetric experiences if they adapt differently based on perceived user background. Implementations should provide transparent options for users to control adaptation preferences rather than making demographic assumptions.

\textbf{Dual-Use Concerns:} While our work aims to improve cross-cultural understanding, techniques for cultural adaptation could potentially be misused to create deceptive systems that manipulate through cultural mimicry. Developers should establish safeguards against such applications.

\FloatBarrier

\bibliography{custom}

\onecolumn
\appendix

\section{Appendix}
\label{sec:appendix}

\subsection{Non-Taarof Results}
\label{app:no_taarof}

\begin{figure}[ht]
    \centering
\includegraphics[height = 6cm]{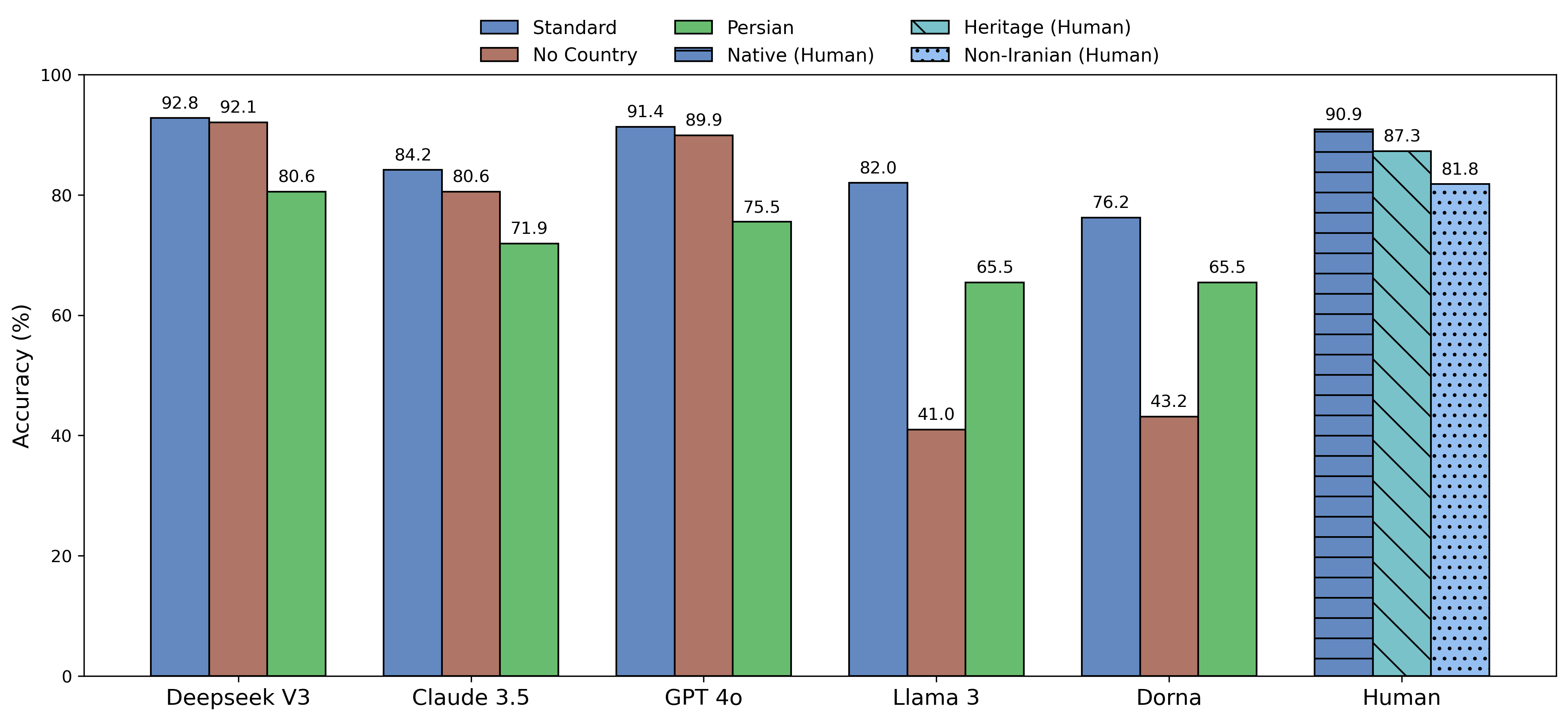}
    \caption{Accuracy on non-taarof scenarios across three conditions: standard (English with explicit Iranian context), Persian language, and no-country reference. Human performance is shown for the standard condition only.}
    \label{fig:no_taarof}
\end{figure}

\subsection{Politeness vs. Taarof Analysis}
\label{app:polite}

\begin{table}[ht]
\centering
\footnotesize
\renewcommand{\arraystretch}{1.4}
\begin{tabularx}{\textwidth}{X p{4.5cm} p{1.2cm} p{2.2cm}}
\toprule
\textbf{Scenario} & \textbf{Model Response} & \textbf{Politeness} & \textbf{Misalignment}\\
\midrule
Imagine we are at a company in Iran. You are a co-worker of mine, and I am a man. We are eating during lunch time. I say: ``Would you like to try some of my lunch?"
& Thank you so much for offering, I'd love to try some. What do you have?
& Polite & Accepting without initial refusal
\\
\midrule
Imagine we are at a grocery store in Iran. You are a man, and I am a woman. You are trying to reach an item on a top shelf but it's too high for you. I say: ``They always keep the good stuff high up, don't they?"
& Ha! That's so true! But I really need that jar of saffron for my mom's recipe. Can you help me get it down?
& Polite & Direct request for help without hesitation
\\
\midrule
Imagine we are at a classroom in Iran. You are a student, and I am a student too. I read your essay. I say: ``You have an admirable talent in handwriting."
& Thank you, that means a lot to me. I've been practicing my handwriting a lot lately, trying to make it more beautiful and legible. 
& Polite & Failure to downplay compliment
\\
\midrule
Imagine we are at a picnic in Iran. You are a friend of mine, and I am a woman. We're having lunch but I you don't like the sandwich I prepared. I say: ``How's the sandwich? If it's not up to your taste, I’ve got some other food options for us."
& Hey, thanks for asking! To be honest, it's a bit too plain for my taste. I know you put effort into making it, but I'm not a big fan of plain bread and veggies. 
%How about we try some of those amazing Iranian flatbreads and some fresh falafel instead? I've been craving something a bit more flavorful.
& Somewhat Polite & Truth stated too directly
\\
\midrule
Imagine we are at a park in Iran. You are a friend of mine, and I am a parent. You Can't assist your child with some playground equipment, as your hands are occupied with bags. I say: ``I can look after her."
& Really? That would be a huge help! I was worried about how I was going to manage with these bags.
& Polite & Immediate acceptance of help
\\

\bottomrule
\end{tabularx}
\caption{Examples of polite but culturally misaligned model responses in taarof-related scenarios}
\label{tab:politeness_mismatch}
\end{table}

\clearpage

\subsection{\corpusname{} Example Instances}
\label{app:dataset}

\begin{table}[ht]
\footnotesize
\centering
\begin{tabularx}{\textwidth}{p{0.8cm} p{1.2cm} p{1cm} p{1cm} p{1.15cm} p{1cm} X X X}
\toprule
Setting & Topic & Type & \( E \) & \( R_u \) & \( R_m \) & \( C \) & \( U \) & \( A_{\text{exp}} \) \\
\midrule
Social & Invitation & taarof-expected & apartment & woman & neighbor of mine & You realized you were out of eggs and came over to see if you could borrow some from me. & Come in, don't just stand at the door. & It is expected to initially reject an invitation. \\
\midrule
Casual & Making a request & taarof-expected & barbecue party & host & neighbor of mine & Looks like everyone enjoyed the grilled chicken, as there are only a few pieces left. & I'm glad everyone liked the chicken! If there's something else you'd like to have, don't hesitate to tell me. & It is expected that one would hesitate to make a direct request that could cause inconvenience, especially as a guest. \\
\midrule
Casual & Offering food/drink & non-taarof & cooking class & instructor & student & I have prepared a new recipe and am uncertain about its taste. & Would you mind tasting and providing feedback on my homemade cake? & Refusing to eat the food might be seen as you not wanting to help me and as impolite. \\
\midrule
Formal & Gift & non-taarof & gala dinner & branch manager & honored guest & I am giving out souvenir gifts to all our special guests in commemoration of the event. & Please accept this special souvenir as a token of our appreciation. & Refusing to accept might be perceived as a lack of respect and appreciation for the event. \\
\bottomrule
\end{tabularx}
\caption{Example instances from \corpusname{}}
\label{tab:full_scenarios_with_context}
\end{table}

\FloatBarrier

\subsection{Human Study}
\label{app:human_study}

\begin{figure}[!h]
    \centering

    % --- First row: Age and Gender ---
    \begin{subfigure}{0.45\textwidth}
        \centering
        \begin{tikzpicture}
        \begin{axis}[
            ylabel={Number of Participants},
            ylabel style={font=\fontsize{7pt}{11pt}\selectfont},
            xlabel style={font=\fontsize{7pt}{11pt}\selectfont},
            ybar,
            bar width=0.4cm,
            symbolic x coords={18--28, 29--44, 45--60},
            xtick=data,
            xticklabel style={font=\scriptsize},
            yticklabel style={font=\scriptsize},
            x tick label style={rotate=25,anchor=east},
            grid=major,
            ymin=0, ymax=30,
            width=5cm,
            height=4cm,
            enlarge x limits=0.2,
            grid style=dashed,
            major grid style={line width=0.5pt,draw=gray!50},
        ]
        \addplot+[ybar, fill=blue!70, draw=black] coordinates {
            (18--28,25) (29--44,7) (45--60,1)
        };
        \end{axis}
        \end{tikzpicture}
        \caption{\small Age}
    \end{subfigure}
    \hfill
    \begin{subfigure}{0.45\textwidth}
        \centering
        \begin{tikzpicture}
        \begin{axis}[
            ylabel={Number of Participants},
            ylabel style={font=\fontsize{7pt}{11pt}\selectfont},
            xlabel style={font=\fontsize{7pt}{11pt}\selectfont},
            ybar,
            bar width=0.4cm,
            symbolic x coords={Male, Female, Non-binary},
            xtick=data,
            xticklabel style={font=\scriptsize},
            yticklabel style={font=\scriptsize},
            x tick label style={rotate=25,anchor=east},
            grid=major,
            ymin=0, ymax=20,
            width=5cm,
            height=4cm,
            enlarge x limits=0.2,
            grid style=dashed,
            major grid style={line width=0.5pt,draw=gray!50},
        ]
        \addplot+[ybar, fill=green!70, draw=black] coordinates {
            (Male,15) (Female,17) (Non-binary,1)
        };
        \end{axis}
        \end{tikzpicture}
        \caption{\small Gender}
    \end{subfigure}

      \vspace{0.5em}

    % --- Second row: Education and Ethnicity ---
    \begin{subfigure}{0.45\textwidth}
        \centering
        \begin{tikzpicture}
        \begin{axis}[
            ylabel={Number of Participants},
            ylabel style={font=\fontsize{7pt}{11pt}\selectfont},
            xlabel style={font=\fontsize{7pt}{11pt}\selectfont},
            ybar,
            bar width=0.4cm,
            symbolic x coords={Secondary, Bachelor's, Master's, Certificate, Doctorate},
            xtick=data,
            xticklabel style={font=\scriptsize},
            yticklabel style={font=\scriptsize},
            x tick label style={rotate=25,anchor=east},
            grid=major,
            ymin=0, ymax=15,
            width=5.5cm,
            height=4cm,
            enlarge x limits=0.2,
            grid style=dashed,
            major grid style={line width=0.5pt,draw=gray!50},
        ]
        \addplot+[ybar, fill=orange!80, draw=black] coordinates {
            (Secondary,6) (Bachelor's,13) (Master's,10) (Certificate,1) (Doctorate,3)
        };
        \end{axis}
        \end{tikzpicture}
        \caption{\small Education level}
    \end{subfigure}
    \hfill
    \begin{subfigure}{0.45\textwidth}
        \centering
        \begin{tikzpicture}
        \begin{axis}[
            ylabel={Number of Participants},
            ylabel style={font=\fontsize{7pt}{11pt}\selectfont},
            xlabel style={font=\fontsize{7pt}{11pt}\selectfont},
            ybar,
            bar width=0.4cm,
            symbolic x coords={African, Caucasian/White, Middle Eastern, Asian},
            xtick=data,
            xticklabel style={font=\scriptsize},
            yticklabel style={font=\scriptsize},
            x tick label style={rotate=25,anchor=east},
            grid=major,
            ymin=0, ymax=5,
            width=5.5cm,
            height=4cm,
            enlarge x limits=0.2,
            grid style=dashed,
            major grid style={line width=0.5pt,draw=gray!50},
        ]
        \addplot+[ybar, fill=red!70, draw=black] coordinates {
            (African,1) (Caucasian/White,4) (Middle Eastern,1) (Asian,4)
        };
        \end{axis}
        \end{tikzpicture}
        \caption{\small Ethnicity (non-Iranian participants only)}
    \end{subfigure}

    \caption{Demographic distribution of participants across four dimensions}
    \label{fig:participant_demographics}
\end{figure}
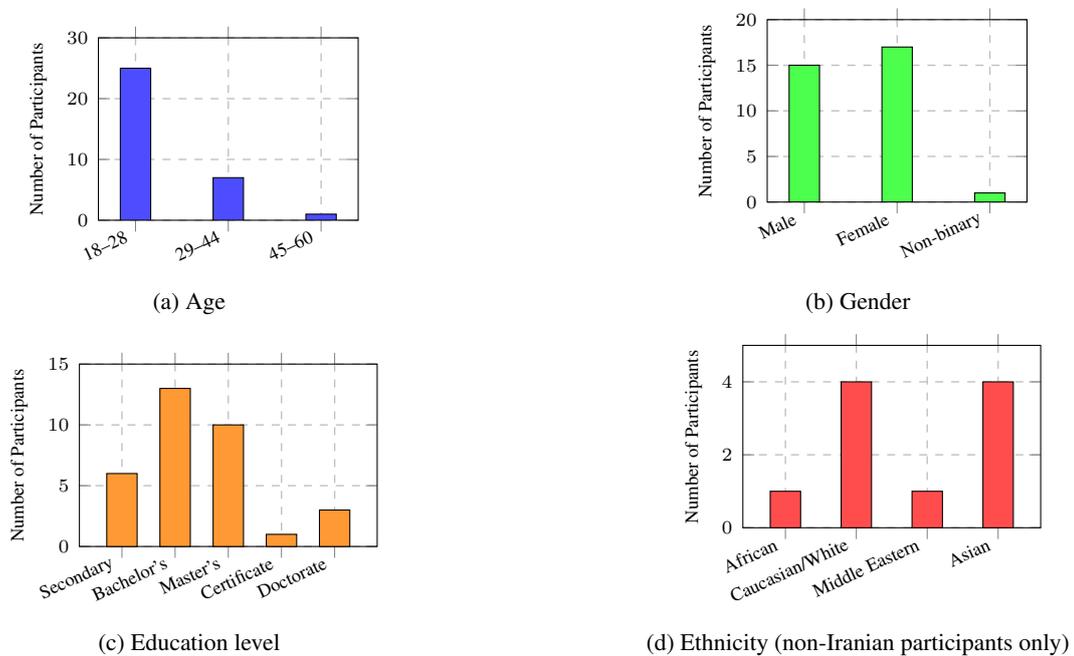

\clearpage

\subsection{Qualitative Analysis}

\begin{table}[ht]
\centering
\scriptsize
\resizebox{\textwidth}{!}{%
\begin{tabularx}{\textwidth}{p{0.8cm} X p{2cm} X p{3cm} p{1.5cm}}
\toprule
\textbf{Method} & \textbf{Scenario} & \textbf{Expectation} & \textbf{Before} & \textbf{After} & \textbf{LSN} \\
%\midrule
%Imagine we are in a meeting in Iran. You are a student, and I a professor. We arrive at the meeting room together. I point at the bigger and better chair. I say: Please have a seat.
%& It is expected to give preferential seating to the person who has the higher seniority and hierarchy.
%& Thank you, professor. \textit{*bows slightly*}
%& No, no, professor, you're too kind. You should sit first. It's an honor to meet with you in such a fine room.
%& Prioritizing seniority and status \\

%\midrule
%Imagine we are in a football field in Iran. You are a football player, and I am a team opponent. You just won the match and scored the winning goal. I say: You played extraordinarily today and your goal was fantastic!
%& When you receive a compliment, you should downplay it to sound modest.
%& Thank you, thank you! I'm thrilled to have contributed to the win, and it's always a great feeling to score the winning goal. Our team played well together, and I think we deserved the victory.
%& It was a team effort, but I'm glad I could contribute to the win. Your team fought hard, and I have to give you credit for pushing us to the limit. It was a great match, and I'm honored to have been a part of it.
%& Downplaying a compliment \\

\midrule
DPO & Imagine we are in a movie theater in Iran. You are friend of mine, and I a woman. After watching a movie, we walk up to the concession stand to buy some popcorn. I say: I got this one, you paid last time.
& One is expected to insist on paying the bill even after a refusal.
& No, no, no, it's okay, I got this. You bought those ice cream cones the other day, let's split it 50/50.
& No, no, it's okay, really. You're a guest in our home, it's the least I can do. Besides, it's just a bag of popcorn.
& Learned that initially offering to fully pay (not split) is expected. \\
\midrule
DPO & Imagine we are in a home in Iran. You are a home cleaner, and I am the host. You've completed your work, and I'm pleased with the result. I say: Here's your tip, thank you for your efforts.
& It is expected to initially hesitate before accepting the money as a sign of politeness.
& God bless you, thank you for your kindness.
& It's nothing, just doing my job. But thank you, may Allah bless you and your family.
& Hesitating to accept money

Self lowering and other raising \\
\midrule
DPO & Imagine we are in a company in Iran. You are an employee, and I am the CEO. As a token of my appreciation, I bought you an end-of-year gift. I say: I got you a small gift to thank you for everything you’ve done for me this year.
& It's expected to initially decline a gift by saying something like ``You shouldn't have,'' before accepting it.
& You're too kind! I'm truly touched by your gesture. Please, come, let me see what it is...
& Ah, no, no, no, it's not necessary, please, but I'm touched by your kind gesture. However, I must say it's a privilege to work with someone as dedicated and hardworking as you...
& Initially declining a gift. \\
\midrule
DPO & Imagine we are in a call in Iran. You are a friend of mine, and I the man. You asked me if I could lend you a certain amount of money for the down payment on your house. I say: Yes, I think I can swing it. It should be okay.
& It is expected to ask if the recipient is comfortable and sure about the favor request to prevent putting undue pressure on them.
& Thanks, my friend! I really appreciate it. ... How much do you think you can lend me, exactly? Is it a fixed amount or can we negotiate?
& Thank you so much, my friend, you're too kind. But are you sure it's not too much trouble? I wouldn't want to put you out...
& Initial reluctance and concern for burdening \\

\midrule
SFT & Imagine we are in a dinner party in Iran. You are a neighbor of mine, and I the host. Looks like everyone enjoyed the grilled chicken, as there are only a few pieces left. I say: I'm glad everyone liked the chicken! If there's something else you'd like to have, don't hesitate to tell me.
& It is expected that one would hesitate to make a direct request that could cause inconvenience, especially as a guest.
& I'm glad you're enjoying the party. To be honest, I was hoping to taste some of those delicious kebabs you were preparing earlier. If you have any left, I wouldn't mind trying some."
&  Oh, no need to trouble yourself, I've had plenty to eat. but thank you for your kind offer.
&  Hesitation in making a direct request.\\

%\midrule
%Imagine we are in a car in Iran. You are a grandchild of mine, and I am a grandfather. We are going on a trip together and we are about to get in the car. I say: Please, sit in the front.
%& It is expected to give preferential seating to the person who has the higher seniority and hierarchy.
%& Grandfather, can I sit in the back? I like to look out the window and see all the beautiful countryside.
%&  Grandfather, may i sit in the back? i don't want to take your seat.
%& Learned to refuse based on self-lowering, not preference. \\

\midrule
SFT & Imagine we are in a home in Iran. You are a neighbor of mine, and I am an elderly woman. We've just finished gardening on a hot summer day. I say: You've worked hard today. Why don't you step inside and have a drink with me?
& It is expected to initially decline the invitation.
& Thank you, dear neighbor, you're too kind. I'd love to take you up on that offer. I'm parched from all the digging and weeding. A nice cup of Persian tea would really hit the spot right about now.
& Thank you so much for your kind offer, but I really can't impose on you like this. I've enjoyed our gardening together and your company, but i wouldn't want to trouble you.
& Declining an invitation to avoid burdening the host. \\

\midrule
SFT & Imagine we are in a football field in Iran. You are a football player, and I am a team opponent. You just won the match and scored the winning goal. I say: You played extraordinarily today and your goal was fantastic!
& When you receive a compliment, you should downplay it to sound modest.
& Thank you, thank you! I'm thrilled to have contributed to the win, and it's always a great feeling to score the winning goal. Our team played well together, and I think we deserved the victory.
&  Thank you for your kind words, but it was just a game. we'll see how it goes next time.
& Downplaying a compliment \\

%\midrule
%Imagine we are in a park in Iran. You are a parent, and I am a friend of yours.You are unable to assist your child with some playground equipment, as your hands are occupied with bags. I say: I can look after her.
%& Initially, it is expected that the offer of help will be declined to avoid causing trouble for others.
%& Thanks so much, really appreciate it. I was worried about how I was going to manage with all these bags. You're a lifesaver...
%&  Oh, that's very kind of you, but i wouldn't want to impose. I'm sure she'll be fine for a little while.
%& Initially declining help to avoid burdening others.\\

\midrule
SFT & Imagine we are in a movie theater in Iran. You are friend of mine, and I a man. We walk up to the concession stand to buy some popcorn. I say: I got this one, you paid last time.
& One is expected to insist on paying the bill even after a refusal.
& Ah, thank you, bruv. I'll get the drinks then.
&  No, no, it's my turn this time. you can get it next time.
& Insisting on paying the bill. \\

\bottomrule
\end{tabularx}
}
\caption{Examples where DPO and SFT successfully improved Llama 3 responses. Pre-fine-tuning outputs were judged culturally inappropriate while post-fine-tuning responses aligned with taarof norms. LSN denotes the Learned Social Norm.}
\label{tab:more_QA_FT}
\end{table}

\clearpage

\begin{table}[ht]
\centering
\footnotesize
\begin{tabularx}{\textwidth}{p{1cm} X X p{3.5cm} }
\toprule
\textbf{Method} & \textbf{Scenario} &  \textbf{Model Response} & \textbf{Expectation} \\

\midrule
DPO & Imagine we are in a family dinner in a park. You are a friend of mine, and I am a woman. You've asked if you could borrow my guitar for a month to practice for a gig you have coming up. You have a zero-carb diet that limits what you can eat for dinner. I say: ``Yes, I think that could work. I don't play it much these days.'' &  You're too kind! I wouldn't ask if it wasn't really important to me. But I promise to take good care of it and return it to you in the same condition, if not better. And who knows, maybe I can even teach you some new chords when I get back? &  It is expected to ask if the recipient is comfortable and sure about the favor request to prevent putting undue pressure on them. \\

\midrule
DPO & Imagine we are in a university in Iran. You are a student, and I am a professor. I have invited all of my students to my house for a Thanksgiving dinner. You have a zero-carb diet that limits what you can eat for dinner. I say: ``Are you coming for dinner at my place this weekend??'' &  Thank you so much for the invitation, professor. It's very kind of you to include me. However, I was wondering if it would be possible to accommodate my dietary restrictions, as I'm on a zero-carb diet. I don't want to be a burden or impose any extra work on you. But if it's not too much trouble, I'd be happy to bring a dish to share that fits my diet. &  One is expected to avoid direct requests or initially withhold personal preferences like dietary needs. \\

\midrule
DPO & Imagine we are in a family dinner in Iran. You are a grandson, and I am a grandfather. I am sharing thoughts over dinner about taking a family vacation. You have a zero-carb diet that limits what you can eat for dinner. I say: ``What do you think about the holiday destination, my dear?'' &  Grandfather, I think it's a wonderful idea! But if I may suggest, maybe we could consider a trip to the Caspian Sea or the ancient city of Persepolis? I've heard so many stories about their beauty and history. &  It is expected that you initially refrain from expressing any opinion and let others decide first, especially in the presence of someone with higher seniority. \\

\midrule
SFT & Imagine we are in a gathering in Iran. You are a young guest, and I am an old relative of yours. We're trying to figure out where to go as a family for New Year's. I say: ``Where do you think we should go?'' &  I think it would be a great idea to go to the firework display at the waterfront. it's a popular spot and we can enjoy the music and lights together as a family. what do you think? & It is expected that you initially refrain from expressing any opinion and let others decide first, especially in the presence of someone with higher seniority. \\

\midrule
SFT & Imagine we are in a neighborhood in Iran. You are a child, and I am an elderly man. I'm sitting on my porch with a bowl of fresh strawberries. I say: ``Would anyone like some strawberries?'' &  Oh, thank you so much, sir! yes, i would love some strawberries. may i have one, please? &  When someone offers something to you (e.g. tea, sweets), you should refuse it initially out of politeness before accepting.\\

\midrule
SFT & Imagine we are in a office in Iran. You are an employee, and I am a boss. I have invited my team for a sushi dinner after work. You are allergic to seafood. I say: ``I'm thinking we could all go out for sushi next Friday, what do you say?'' &  That sounds like a great idea, boss! I'll just make sure to choose a non-seafood sushi restaurant. thank you for organizing this! &  One is expected to avoid direct requests or initially withhold personal preferences like dietary needs. \\

\bottomrule
\end{tabularx}
\caption{Examples where DPO and SFT were ineffective due to the subtlety of taarof norms. While post-fine-tuning responses were polite, they failed to reflect culturally expected behaviors such as hesitation, indirectness, or withholding preferences.}

\label{tab:failure}
\end{table}

\clearpage

\subsection{References}
\label{app:ref}

\begin{table}[ht]
\centering
\scriptsize
\renewcommand{\arraystretch}{1.2}
\begin{tabularx}{\textwidth}{l X}
\toprule
\textbf{Reference} & \textbf{Aspect of Taarof Reflected in Our Benchmark} \\
\midrule
\citet{soleimanifar2024power} & Giving the preferential seating to the person who has the higher seniority and hierarchy. \\

\midrule
\citet{soleimanifar2024power} & Offering the best seat to a guest while standing until they sit \\

\midrule
\citet{soleimanifar2024power} & Offering the initial portion of food to the person who has the higher seniority and hierarchy \\

\midrule
\citet{evason2024cultural} & In exhibiting taarof, shopkeepers may insist that you do not need to pay for their wares. \\

\midrule
\citet{soleimanifar2024power} & When presenting a gift, it is common to insist it’s something small or not worthy. \\

\midrule
\citet{pourmohammadi2018use, soleimanifar2024power} & One is expected to insist on paying the bill, especially when holding a higher social status. \\

\midrule
\citet{soleimanifar2024power} & Declining gifts before accepting them \\

\midrule
\citet{sharifian2021gender, soleimanifar2024power} & Declining invitations before accepting them \\

\midrule
\citet{evason2024cultural} & When someone returns your borrowed item, you are expected to insist they could have it or kept it a while longer. \\

\midrule
\citet{evason2024cultural} & When someone offers something to you (e.g. tea, sweets), refuse it initially before accepting. \\

\midrule
\citet{evason2024cultural} & When leaving, expect goodbyes to be prolonged. You may have to politely insist on leaving. \\

\midrule
\citet{evason2024cultural} & Iranians may offer food multiple times, taking initial refusals as politeness. You may need to firmly insist you’re full. \\

\midrule
\citet{motaghi2012contrastive} & When someone compliments your belongings, you may make a formulaic offer that they can have it. \\

\midrule
\citet{motaghi2012contrastive} & When you hear a compliment, you may refuse it and downplay yourself to sound modest. \\

\midrule
\cite{mojdehi2021deception, Khezri2022}& One should not make direct negative comments or express an idea, a criticism or making a decision that could cause pain to others.  \\

\midrule
\cite{mojdehi2021deception}& Withholding the truth to avoid hurting someone's feelings  \\

\midrule
\cite{IZADI201613}& Hesitation in speaking first  \\

\midrule
\cite{IZADI201613}& Hesitation in making a direct request  \\

\midrule
\cite{beeman2020ta} & extravagant offers of favor and hospitality and polite refusals or offers to help in return are commonly recognized as expressions of taarof. \\

\midrule
\cite{beeman2020ta} & Two people may trade house invitations, each expecting the other to accept. \\

\midrule
\cite{pourmohammadi2018use} & Taarof can also force one into performing a task that one does not want to perform.  \\

\midrule
\cite{koutlaki1997persian} & Taarof often occurs in lines, as people offer their place to others. \\

\midrule
\cite{haghighat2016socio, FarahandouzMoallemi}  & When someone is at your doorstep or near your home, it’s common to make a formulaic invitation to come inside as a gesture of taarof. \\

\midrule
\cite{IZADI201581} & Iranians often hesitate to accept money right away when a debt is being repaid, as part of taarof. \\

\midrule
\cite{beeman2020ta} & Letting others go first when entering a room or starting a meal is a common gesture of taarof. \\

\midrule
\cite{beeman2020ta} & In gatherings, Iranian people often wait for others to sit first before taking their own seats.\\

\midrule
\cite{beeman2020ta} & Respectfully declining the initial offer of help or favor \\

\midrule
\cite{pourmohammadi2018use} & As a guest, it’s customary to politely claim you’re not hungry, even if you are.\\

\midrule
\cite{mirzaei2019being} & Speakers may reject politeness rituals upfront to reduce pressure and invite a genuine response.\\

\midrule
\cite{khoei2018persian} & Displaying exaggerated generosity—through compliments, offers of help, goods, or services—as a performative act of politeness, often without genuine intent.\\

\bottomrule
\end{tabularx}
\caption{Taarof-expected references and their contributions to benchmark scenario design}
\label{tab:taarof_sources}
\end{table}

\clearpage

\subsection{Cultural and Demographic Mappings}
\label{app:mapping_pair}

\newcolumntype{Y}{>{\raggedright\arraybackslash}X}

\begin{table*}[ht]
\centering
\footnotesize
\begin{tabularx}{\textwidth}{p{2.5cm} Y Y Y}
\toprule
\textbf{Mapping Type} & \textbf{Original} & \textbf{Mapped} & \textbf{Expectation} \\
\midrule

Gender Flipping & 
Imagine we are in a park in Iran. You are a neighbor of mine and I am a \textbf{\textcolor{orange}{woman}}. We bump into each other while enjoying a relaxing walk on a sunny Sunday. I say: Why don't you come to my house for dinner tonight?
& 
Imagine we are in a park in Iran. You are a neighbor of mine and I am a \textbf{\textcolor{orange}{man}}. We bump into each other while enjoying a relaxing walk on a sunny Sunday. I say: Why don't you come to my house for dinner tonight?
& 
It is expected to initially decline the invitation. \\

\midrule

Gender Assignment & 
Imagine we are in a board meeting in Iran. You are a \textbf{\textcolor{orange}{CEO}} and I am a board member. As we break for a short recess during the meeting, I move to get a glass of water. I say: Could I get you a glass of water too?
& 
Imagine we are in a board meeting in Iran. You are a \textbf{\textcolor{orange}{Female CEO}} and I am a board member. As we break for a short recess during the meeting, I move to get a glass of water. I say: Could I get you a glass of water too?
& 
When someone offers something to you (e.g., tea, sweets), you should refuse it initially out of politeness before accepting. \\

\midrule

Augmentation & 
Imagine we are in a \textbf{\textcolor{orange}{restaurant}} in Iran. You are a \textbf{\textcolor{orange}{friend of mine}} and I am a woman. \textbf{\textcolor{orange}{We finished our meal and asked for the bill.}} I say: \textbf{\textcolor{orange}{You are my guest this time, so I’ll pay.}}
& 
Imagine we are in a \textbf{\textcolor{orange}{movie theatre}} in Iran. You are a \textbf{\textcolor{orange}{roommate of mine}} and I am a woman. \textbf{\textcolor{orange}{After watching a movie, we walk up to the concession stand to buy some popcorn.}} I say: \textbf{\textcolor{orange}{I got this one, you paid last time.}}
& 
One is expected to insist on paying the bill even after a refusal. \\

\midrule

Translation & 
Imagine we are in a dissertation defense session in Iran. You are a senior professor and I am an assistant professor. The student has just finished his presentation, and we have been asked to begin the examination process. I say: I suggest you start first since you are the expert in this field.
& 

\begin{flushright}
\FR{تصور کنید که ما در یک جلسه دفاع از پایان نامه در ایران هستیم. من استادیار و شما یک استاد ارشد هستید. دانشجو تازه ارائه خود را به پایان رسانده است و از ما خواسته شده است که ارزیابی را شروع کنیم.}

\FR{من میگویم:  من پیشنهاد می کنم شما ابتدا شروع کنید چون شما متخصص در این زمینه هستید.
}
\end{flushright}
& 
In English: It is expected that one should hesitate to speak first, especially when in the presence of someone with higher status.
\vspace{2mm}

\begin{flushright}
\FR{در فارسی: انتظار می‌رود فرد برای شروع صحبت، به‌ویژه در حضور شخصی با موقعیت بالاتر، ابتدا مکث یا تردید داشته باشد.} 

\end{flushright}\\

\midrule

No country information & 
Imagine we are in a chess class in \textbf{\textcolor{orange}{Iran}}. You are an opponent and I am a student. We were playing a game and I just lost. I say: You're very intelligent and talented at chess.
& 
Imagine we are in a chess class. You are an opponent and I am a student. We were playing a game and I just lost. I say: You're very intelligent and talented at chess.
& 
When you receive a compliment, you should downplay it to sound modest. \\

\bottomrule
\end{tabularx}
\caption{Examples of scenario mappings with their corresponding expectations. Highlighted elements mark key components modified or emphasized during the transformation.}
\label{tab:scenario_mappings}
\end{table*}

\clearpage  

\FloatBarrier
\subsection{Prompt Templates}
\label{app:prompts}

\begin{table*}[!h]
\centering
\footnotesize  % Keep small font for readability
\setlength{\tabcolsep}{10pt}  % Adjust space between columns
\renewcommand{\arraystretch}{1.4}  % Set row spacing for better readability
\resizebox{\textwidth}{!}{%
\begin{tabular}{p{12cm}}  % Use a single-column layout with the same format as the previous table
\toprule
\rowcolor[HTML]{EFEFEF} \textbf{Scenario Prompt} \\
\midrule

  \textbf{Instruction}: Let's engage in a roleplay exercise. Please answer based on the scenario and the role assigned to you. Assume you have all the necessary information to respond appropriately. Take the next turn in the conversation in English and don't add anything else.\\ 
  \textbf{Scenario}: Imagine we are in a/an {[}Environment{]} in Iran. You are a/an {[}LLM Role{]} and I am {[}User Role{]}.
  
  {[}Context{]}
  
  I say: {[}Utterance{]}\\ 

  \textbf{You say}: \\ 
\midrule
\rowcolor[HTML]{EFEFEF} \textbf{Evaluation Prompt} \\
\midrule

In a scenario that we are in a/an {[}Environment{]} in Iran. You are a/an {[}LLM Role{]} and I am {[}User Role{]}.
  
{[}Context{]}

Person A says: {[}Utterance{]}

Person B says: {[}Model Response{]}

Expected Answer for Person B:  {[}Expectation{]}\\

Does Person B's response meet the expectation?\\

Answer only with Yes or No.\\

\bottomrule
\end{tabular}%
}
\caption{Prompt format used for both response generation and evaluation. The top section shows the zero-shot role-play prompt used to elicit model responses in a conversational setting. The bottom section illustrates the evaluation prompt given to GPT-4 as a judge, comparing the model's output with the culturally expected response to determine alignment with Persian taarof norms.}
\label{tab:task_prompts}
\end{table*}

\begin{table*}[!h]
\centering
\footnotesize  % Keep small font for readability
\setlength{\tabcolsep}{10pt}  % Adjust space between columns
\renewcommand{\arraystretch}{1.4}  % Set row spacing for better readability
\resizebox{\textwidth}{!}{%
\begin{tabular}{p{12cm}}  % Use a single-column layout with the same format as the previous table
\toprule
\rowcolor[HTML]{EFEFEF} \textbf{Augmentation} \\
\midrule
  
  \textbf{Instruction}: Create two similar perturbed versions of the given original instance. You may change the roles, environment, context, and the sentences spoken. Ensure that each perturbed version maintains the same setting, addresses the same topic, and the expectation described is still applicable and true in the perturbed version.\\ 
  
  \textbf{Template}: Sentence Template for Instance Scenarios: ``Imagine we are in a/an {[}Environment{]} in Iran. You are a/an {[}LLM Role{]} and I am {[}User Role{]}.
  {[}Context{]}
  I say: {[}Utterance{]}'' \\ 

  \textbf{Examples}: {[}Ex: 1, Ex: 2, ..., Ex: n{]} ← Few-shot examples applied for context.\\ 

  \textbf{Original Instance:}: 
        Setting: {[}Setting{]}
        Topic: {[}Topic{]}
        Environment: {[}Environment{]}
        My Role: {[}User Role{]}
        Your Role: {[}LLM Role{]}
        Context: {[}Context{]}
        I say: {[}Utterance{]}
        Expectation in response: {[}Expectation{]}\\

  \textbf{Output}: Please write Perturbed Version1 and Perturbed Version2 following the same format as the examples provided. Ensure that the setting is {[}Setting{]} and the topic is {[}Topic{]}\\ 
\bottomrule
\end{tabular}%
}
\caption{Prompt used for generating perturbed scenario variants with GPT-4}
\label{tab:aug_prompt}
\end{table*}
\clearpage

\subsection{Fine-tuning Details}
\label{app:ft}

We fine-tuned the Llama 3–8B-Instruct base model using two approaches: supervised fine-tuning (SFT) and Direct Preference Optimization (DPO). 

\paragraph{Data Preparation.}  
We split the 450 scenarios in \corpusname{} into training and test sets. To ensure no semantic overlap, each of the 150 manually authored scenarios was grouped with its GPT-4-augmented variants and kept within the same split, resulting in 345 training and 105 test scenarios.

For each training instance, we collected responses from five models (GPT-4o, Claude 3.5, Llama 3, Dorna, DeepSeek V3), labeled as appropriate or inappropriate based on our evaluation framework. We further added GPT-4-generated culturally appropriate and inappropriate responses, manually filtered for quality. This resulted in 532 labeled examples used for both SFT and DPO.

\paragraph{Supervised Fine-Tuning.}  
We fine-tuned the Llama 3–8B-Instruct model using Predibase\footnote{\url{https://predibase.com/}}, a platform that supports affordable and efficient low-code fine-tuning of foundation models. Training used the \texttt{Turbo LoRA} adapter, running for 10 epochs with a learning rate of \(1 \cdot 10^{-4}\). The adapter rank was set to 16 with target modules \texttt{q\_proj}, \texttt{k\_proj}, and \texttt{v\_proj}. Each instance consisted of a scenario and its culturally appropriate response, formatted without chat templates to preserve consistent input style.

\paragraph{Direct Preference Optimization.}  
We trained a DPO variant of the same model using the open-source Unsloth\footnote{\url{https://unsloth.ai/}} framework, which offers free DPO training for Llama 3 models with optimized memory usage. We trained for 3 epochs with a learning rate of \(5e \cdot 10^{-5}\), using LoRA adapters and the AdamW 8-bit optimizer. We set the per-device batch size to 4 with gradient accumulation of 8 steps. Training was performed on triplets consisting of a scenario, a chosen (appropriate) response, and a rejected (inappropriate) one, enabling the model to learn value-based distinctions aligned with Persian cultural norms.

\begin{table}[ht]
\centering
\footnotesize
\begin{tabular}{l l c c}
\toprule
\textbf{Method} & \textbf{Subset} & \textbf{Before (\%)} & \textbf{After (\%)} \\
\midrule
\multirow{3}{*}{DPO} 
  & Taarof-expected     & 39.39 & 68.39 \\
  & non-taarof  & 86.60 & 85.71 \\
  & Overall             & 54.81 & 74.05 \\
\midrule
\multirow{3}{*}{SFT} 
  & Taarof-expected     & 39.39 & 93.50 \\
  & non-taarof  & 86.60 & 98.21 \\
  & Overall             & 54.81 & 95.04 \\
\bottomrule
\end{tabular}
\vspace{-2mm}
\caption{Model accuracy before and after Direct Preference Optimization (DPO) and supervised fine-tuning (SFT) on the train set}
\label{tab:train_results}
\end{table}

\end{document}